\documentclass{ieeeaccess}
\usepackage{graphicx}
\usepackage{xcolor}
\usepackage{multirow,tabularx}
\usepackage[numbers,square]{natbib}
\usepackage{authblk}
\usepackage{colortbl}
\usepackage{float}
\usepackage{graphicx}
\usepackage{caption}
\usepackage{subcaption}
\usepackage{dirtree}
\usepackage{adjustbox}
\usepackage{enumitem}
\usepackage{textgreek}
\usepackage{kantlipsum}

\graphicspath{ {./images/} }
\usepackage[numbers,square]{natbib}
\usepackage{amsmath,amssymb,amsfonts}
\usepackage{algorithmic}
\usepackage{graphicx}

\usepackage{caption,setspace}
\usepackage{textcomp}
\def\BibTeX{{\rm B\kern-.05em{\sc i\kern-.025em b}\kern-.08em
    T\kern-.1667em\lower.7ex\hbox{E}\kern-.125emX}}
    
\begin{document}
\history{Date of publication xxxx 00, 0000, date of current version xxxx 00, 0000.}
\doi{10.1109/ACCESS.2017.DOI}

\title{ Multi Visual Modality Fall Detection Dataset}
\author{\uppercase{Stefan Denkovski}\authorrefmark{1, 2},
\uppercase{Shehroz S. Khan\authorrefmark{1, 2}, Brandon Malamis\authorrefmark{2}, Sae Young Moon\authorrefmark{1, 2}, Bing Ye \authorrefmark{1, 2}, Alex  Mihailidis}.\authorrefmark{1, 2}}
\address[1]{KITE Research Institute, Toronto Rehabilitation Institute – University Health Network, Toronto, ON M5G 2A2, Canada}
\address[2]{Institute of Biomedical Engineering, University of Toronto, Toronto, ON M5G 2A2, Canada}

\tfootnote{This research was support by NCE:AGE-WELL, Fund 499052.}

\markboth
{Author \headeretal: Preparation of Papers for IEEE TRANSACTIONS and JOURNALS}
{Author \headeretal: Preparation of Papers for IEEE TRANSACTIONS and JOURNALS}

\corresp{Corresponding author: Stefan Denkovski (e-mail: stefan.denkovski@mail.utoronto.ca).}

\begin{abstract}
Falls are one of the leading cause of injury-related deaths among the elderly worldwide. Effective detection of falls can reduce the risk of complications and injuries. Fall detection can be performed using wearable devices or ambient sensors; these methods may struggle with user compliance issues or false alarms. Video cameras provide a passive alternative; however, regular RGB cameras are impacted by changing lighting conditions and privacy concerns. From a machine learning perspective, developing an effective fall detection system is challenging because of the rarity and variability of falls. Many existing fall detection datasets lack important real-world considerations, such as varied lighting, continuous activities of daily living (ADLs), and camera placement. The lack of these considerations makes it difficult to develop predictive models that can operate effectively in the real world. To address these limitations, we introduce a novel multi-modality dataset (MUVIM) that contains four visual modalities: infra-red, depth, RGB and thermal cameras. These modalities offer benefits such as obfuscated facial features and improved performance in low-light conditions. We formulated fall detection as an anomaly detection problem, in which a customized spatio-temporal convolutional autoencoder was trained only on ADLs so that a fall would increase the reconstruction error. Our results showed that infra-red cameras provided the highest level of performance (AUC ROC=0.94), followed by thermal (AUC ROC=0.87), depth (AUC ROC=0.86) and RGB (AUC ROC=0.83). This research provides a unique opportunity to analyze the utility of camera modalities in detecting falls in a home setting while balancing performance, passiveness, and privacy.
\end{abstract}

\begin{keywords}
 fall detection, multi-modal, autoencoder, anomaly detection, deep learning, computer vision.
\end{keywords}

\titlepgskip=15pt

\maketitle

\section{Introduction}

Falls are one of the leading causes of injury-related deaths among the elderly worldwide \cite{WHOFalls2021, KramarowEChenLHHedegaardH2015} and they are a major cause of both death and injury in people over 65 years of age \cite{centers2006fatalities, gillespie2012interventions}. The faster an individual receives help after a fall, the lower the risk of complications arising from the fall \cite{Stinchcombe2014, Rubenstein2002, Tinetti1993}. Fall detection systems improve the ability of older adults to live independently and “age in place” by ensuring they receive support when required. However, fall detection is a challenging problem  both from predictive modeling and practical considerations (such as low false alarm rate, privacy) \cite{Igual2013}. 

Predictive modeling challenges faced in fall detection include the rarity and diversity of fall events \cite{Khan2017}. Previous studies by Stone et al.,\cite{stone2014fall} found 454 falls in 3339 days worth of data and Debard et al., \cite{debard2012camera} found 24 in 1440 days worth of data.  Therefore, it is very challenging and consuming to run studies over a long duration, and even then they may still contain few falls which may not be sufficient enough for building robust classifiers \cite{Khan2016}. Another challenge is that fall events last only for short intervals in comparison with normal ADLs \cite{Khan2017}. Finally, each rare fall event can vary greatly from one another, making it difficult to strictly define a well-defined class or to capture all possible variations in a dataset \cite{Khan2016, charfi2012definition, Tran2018}.
ƒ
Many possibilities exist for the practical implementation of fall-detection systems.  However, because solutions are ultimately intended to be implemented in a person's daily life, they should be easy to live with. Thus, passive systems are ideal, as they require no input from the user. They can monitor the environment before detecting a fall. These systems are preferable because a user may be unresponsive after a fall, or they may not be wearing their device \cite{Chaudhuri2014} \cite{Chaudhuri2015}. However, older adults express privacy concerns regarding passive systems such as cameras.  In addition to these concerns, systems must balance high sensitivity to falls while maintaining a low false alarm rate. Missed falls are potentially dangerous to users, who may rely on the system to detect falls. In contrast, a system with a high false alarm rate can cause many issues. If a system automatically calls an ambulance or even notifies loved ones, a high false alarm rate would result in large bills or potentially ignored cases of real falls and eventual rejection of the system.

Therefore, the key factors to consider when designing a system are (i) passive, (ii) privacy protection and, (iii) perform with a high fall detection rate and low false alarm rate. In this paper, we introduce a novel multi-camera and mutli-modal fall detection dataset containing fall and ADL from 30 healthy adults and ADL (with no fall) from 10 older adults. This dataset is collected data in a seminaturalistic setting inside a designed home. The experiments are designed to emulate real-world scenarios and sequences of events. It contains six visual modalities mounted on the ceiling and four additional wearable modalities. The six visual modalities consisted of two infra-red cameras, two depth cameras, a thermal camera and an RGB camera. The four wearable modalities were accelerometer, PPG, GSR and temperature. We did comprehensive experiments on detecting falls using a customized 3D convolutional autoencoder and showed that infra-red modalities performed the best among others, including depth, thermal and RGB cameras.

\subsection{Literature Review}

A wide range of modalities, methods of capturing information, and subsequent systems have been explored for fall detection. They can be divided into various groups, such as wearables, ambient sensors and computer vision-based sensors \cite{Mubashir2013} \cite{Wang2020}. Khan and Hoey \cite{Khan2017}, presented a review of fall detection methods based on the availability of fall data during model training. 

Wearable systems often incorporate accelerometers, gyroscopes, and inertial measurement units (IMUs) to detect falls. However, other modalities may also be used such as an EEG or a barometric Sensor \cite{Martinez-Villasenor2019} \cite{Pierleoni2016}. 
Systems that use accelerometers or IMUs are relatively affordable and accurate \cite{Chaudhuri2014} \cite{Mubashir2013} \cite{Lee2005}. However, these systems are invasive and require the user to constantly wear and charge the device. This can lead to many missed falls as one may not wear the device all the time, for e.g. while charging or bathing. Chaudhuri et al. \cite{Chaudhuri2015} found that the wearable device was not was not worn for two-thirds of falls events. Additionally, older adults may be hesitant to wear a device because of the stigma it may imply regarding independence \cite{Chaudhuri2017}. Despite the benefits of wearables, the real-world short comings of this type of device may limit its successful deployment. 

Ambient or environmental systems, such as pressure mats, radar/Doppler, microphones, and motion sensors, typically use a wide range of sensors within the home to determine the user’s activities \cite{Ma2014} \cite{Martinez-Villasenor2019}. These systems may be difficult to install or have high false-alarm rates owing to environmental noise \cite{Mubashir2013} \cite{Riquelme}.

Computer vision systems cam typically rely on traditional imaging modalities and techniques to determine when a fall has occurred \cite{gutierrez2021comprehensive}.  They often use RGB or depth camera modalities; however, other visual modalities have also been used. These systems are generally passive, low-cost, and can achieve high performance. However, they struggle to maintain user privacy. Imaging modalities, such as depth and thermal cameras, may help alleviate  this issue by obscuring identifiable features. The techniques used to analyze images vary, but 3D bounding boxes and background subtraction are among some of the most popular approaches. Recently, deep learning techniques have been applied to improve results. Modeling techniques include 2D-LSTM models, 3D CNNs. 

Computer vision systems rely have traditionally relied on classical machine learning techniques to determine when a fall has occurred \cite{gutierrez2021comprehensive, ramachandran2020survey}. They often use RGB or depth camera modalities; however, other visual modalities have also been used. These systems are generally low-cost and can achieve high performance. However, they struggle to maintain user privacy and may be limited in the area they are installed. Imaging modalities, such as depth and thermal cameras, may help alleviate this issue by obscuring identifiable features, though a person may still be identified. The techniques used to analyze images vary \cite{ramachandran2020survey}. 3D bounding boxes and background subtraction are among some of the most popular traditional approaches. Recently, deep learning techniques have been applied to improve results. Modeling techniques include CNN-LSTM models and 3D CNNs have shown promising results \cite{gutierrez2021comprehensive}.

We now present a review of some of the fall detection datasets. Our review of fall detection datasets and multi-modal fall detection datasets is limited to those that contain at least one visual modality.

\subsection{Unimodal Datasets}

A review of existing publicly available fall datasets containing only a singular visual modality is outlined below. 

Charfi et al. \cite{charfi2012definition} released an unnamed dataset containing 197 falls and 57 wall mounted videos of normal activities. It comprised four background scenes, varying actors, and different lighting conditions. The directions of activities relative to the camera are varied to reduce any impact that the directional camera location may have. All activities and falls started with a person in the frame and were segmented into short video clips of specific actions. To determine falls, manually generated bounding boxes were used to extract features, and an SVM was used to classify falls.

The multi-cameras fall dataset collected by Auvinet et al. \cite{auvinet2010fall} contained eight IP cameras placed around the room. It included only 24 scenarios, 22 of which contained falls after a short activity. These videos were segmented into clips lasting less than 30 seconds. All rooms were well lit, with minimal furniture placed in the center of the room. The small size of the dataset and other limiting factors could limit the development of generalized models in real-world settings. Methods included background subtraction to obtain a silhouette, in which the vertical volume distribution ratio was used to classify falls. 
 
The Kinect Infrared dataset published by Mastorakes et al., \cite{mastorakis2014fall} used one camera placed at eye level in the middle of a wall.  The dataset included three variations of camera angles (backward, forward and sideways) totaling 48 video falls. The direction of the fall is important because of the eye-level placement of the camera. In addition to falls, activities were performed while sitting, lying  on the floor and “picking up an item from the floor” were performed by eight different participants. Additionally, two participants were asked to perform activities slowly to replicate the movements of older adults. OpenNI and depth data were used from the Kinect sensor to generate a 3D bounding box whose first derivative parameters were analyzed to classify falls.

The EDF/OCCU datasets contain two view points of a kinetic camera \cite{Zhang2014}. Their data were then divided into two datasets. A non-occlusion dataset, EDF, of 40 falls and 30 actions, and an occlusion dataset, OCCU, of 30 occluded falls and 80 actions were used. Both viewpoints were from cameras placed at eye level, and thus 'directional falls' were performed. Room furniture and fall variations were minimal, with most variations related to the direction of falls. Occlusion was introduced by a single bed positioned to block the view of the bottom half of a fall. Actions in the dataset without occlusions included; picking things off the floor, sitting on the floor and lying on the floor. These same actions were performed in the occluded dataset, with the further addition of tying shoelaces. All actions, except for lying on the floor were occluded by the bed.

The SDU dataset comprises ten young adults performing six simple actions captured with a depth camera\cite{Ma2014}. These include falling, bending, squatting, sitting, lying down and walking. These actions were repeated 30 times each under the following conditions: carrying or not carrying a large object, lighting on or off, room layout, camera direction, and position relative to the camera. Despite the large variation in each repetition of the action, the eye-level placement of the camera indicates that the direction is still important. Additionally, all actions were still short and segmented with an average length of eight seconds per clip. 

The KUL simulated fall detection dataset was designed to address many of the shortcomings that these datasets face when real-world factors were considered  \cite{Baldewijns2016}.Five IP cameras, 55 fall scenarios, and 17 ADL videos were used. The dataset includes realistic environments (in terms of furnishings) and longer videos instead of short segmented clips of various activities. Five goals were highlighted to improve the real-world effectiveness of the dataset. This includes realistic settings, fall scenarios, improved balance of ADL to fall activities, including real-world challenges (occlusions, partial falls, lighting..etc.), and continuously recording the data. However, cameras mounted on the walls were used, and no older adults were included in the dataset.

The thermal fall detection dataset was designed to replicate the KUL simulated fall dataset but with a thermal camera \cite{vadivelu2016thermal}. This dataset contains only nine video segments with ADL activities and 35 segments with fall scenarios. However, this dataset only uses a single eye level mounted thermal camera with a slightly more constrained field of view. The same limitations from the KUL simulated fall datasets also apply, as they implemented that datasets set up.

\begin{table*}[]
\caption{Fall Detection Datasets with Unimodal Camera(s). }
\label{table:unimodaldatasets}
\centering
\resizebox{\textwidth}{!}{%
\begin{tabular}{|c|c|c|c|c|c|c|c|}
\hline \textbf{ Datasets} & \textbf{Modalities} & \begin{tabular}[c]{@{}l@{}}\textbf{Number of }\\\textbf{Cameras}\end{tabular} & \begin{tabular}[c]{@{}l@{}}\textbf{Varied} \\\textbf{Lighting}\end{tabular}   & \begin{tabular}[c]{@{}l@{}}\textbf{Contextual} \\\textbf{Activities}\end{tabular}   & \begin{tabular}[c]{@{}l@{}}\textbf{Wall Mounted} \\\textbf{Cameras}\end{tabular}  & \textbf{Occlusions} & \begin{tabular}[c]{@{}l@{}}\textbf{Privacy } \\\textbf{Protecting}\end{tabular}    \\ \hline
Charfi et al. \cite{charfi2012definition}              & RGB        & 1                      & Yes             & No                   & Yes                  & Yes                                 & No                 \\ \hline
Multi Cameras Fall Dataset \cite{auvinet2010fall}            & RGB        & 8                      & Yes              & No                   & Yes                  & Yes                                & No                 \\ \hline
Kinect Infrared \cite{mastorakis2014fall}          & Depth      & 1                      & No              & No                   & Yes                  & No                                & Yes                \\ \hline
EDF/OCCU  \cite{Zhang2014}                  & Depth      & 2                      & No              & No                   & Yes                  & Yes                               & Yes                \\ \hline
SDU \cite{Ma2014}                      & Depth      & 1                      & No              & No                   & Yes                  & Limited                          & Yes                \\ \hline
Thermal Fall \cite{Baldewijns2016}             & Thermal    & 1                      & Yes             & Yes                  & Yes                  & Yes                             & Yes                \\ \hline
KUL Simulated Fall \cite{vadivelu2016thermal}       & RGB        & 5                      & Yes             & Yes                  & Yes                  & Yes                             & No           \\ \hline     
\end{tabular}%
}
\end{table*}

\

Computer vision datasets can vary in several main categories: camera type, location, lighting, occlusions in the scene, and recorded participant activities.

Different camera types or modalities struggle with various considerations. cameras can capture clear images in general lighting conditions; however, they may not work well under poor lighting conditions, such as dark or night time scenarios. RGB cameras also do not offer any level of privacy. Depth cameras, such as the defunct Microsoft Kinect camera, are a popular alternative to an RGB camera because they can provide a light independent image and protect the privacy of individuals \cite{Ma2014, mastorakis2014fall, Martinez-Villasenor2019, Wang2020}. Vision modalities, such as thermal and depth cameras obscure identifying features, while still providing silhouettes of individuals. In addition, they may perform better in certain scenarios such as those with poor lighting or visually busy environments due to independent lighting and silhouette segmentation. 

The location of the camera in the room can also affect the results. The location of the camera in the room can also affect the results. Most of the previous datasets commonly have a camera mounted on a wall at or above eye level, as shown in Table \ref{table:unimodaldatasets}. The problem with this placement is that a fall looks very different depending on its direction (across or in-line with the field-of-view). As such, these datasets include strict definitions of the orientation of falls relative to the camera. This limits the variety of falls to a short list of possible variations \cite{Martinez-Villasenor2019}. In addition to being affected by the fall direction, cameras mounted at eye level are more susceptible to occlusions, blocking the view of the participant and  affecting the cameras' ability to detect falls (e.g. behind furnitures). Some datasets make efforts to include furniture/occlusions in the dataset, however they may be limited to a single chair or bed (The Multiple Camera \cite{auvinet2010fall}, EDF/OCCU \cite{Zhang2014}, SDU Datasets\cite{Ma2014}), while others have no furniture in frame (Mastorakis et al. \cite{mastorakis2014fall}). To mitigate these limitations (i.e., orientation of falls and occlusions) cameras can be placed on the ceiling. This helps provide a similar view of every fall, and removes furniture that may be between the subject and camera \cite{luo2012design} \cite{Gasparrini2014}. 

As with falls, the ADL performed by the participants in the datasets varied widely. Activities are often segmented into very short and specific motions such as a single squat, picking up an object, taking a seat, and lying down..etc. These short and specific videos last from five to 60 seconds may limit a system’s ability to generalize to the real world. This is because of the oversimplification and trivialization of diversity and problems that can arise in activities and falls. In response to the lack of real-world considerations Baldewijns et al. \cite{Baldewijns2016} attempted to create a dataset that considers these factors. However, the dataset still lacks ceiling mounted cameras, varying environments, and older adult participants. A table consisting of these “real” world factors are outlined in Table \ref{table:unimodaldatasets}.


\subsection{Multimodal Datasets}

In recent years, the ability to merge a variety of sensors to improve performance has become of interest in fall detection systems and several multi-modal datasets have emerged. This is because a multi-modal approach provides different sources of information that can help compensate for the deficiencies in each other.

Combining modalities may complement each other from a technical perspective, but it may cause practical consideration to be further impaired. More modalities mean more sensors or cameras, increasing costs and potentially more inconvenience to the user. Because a wide range of modalities are used for fall detection a wide range of combinations is possible. Thus it is important to select modalities that are complimentary to each other without increasing practical costs to the user. These practical or real world considerations for multi-modal datasets are highlighted in Table ~\ref{table:multimodaldatasets}. 

The UP fall detection data set contained two cameras at eye level with frontal and lateral views \cite{Martinez-Villasenor2019}.  Other modalities were captured through five IMU sensors, one EEG headset and six infrared motion sensors placed in a grid. The activities were limited to six simple motions and five fall variations. Activities varied from 10 to 60 seconds and were segmented from each other. The limited fall directions and short segmented activities limit the real-world implications of this dataset. In addition, many of the chosen modalities were also impractical. Five IMU sensors are difficult to implement; selection of only most relevant sensors may be more plausible, which was performed by the authors in a follow up paper \cite{martinez2020design} \cite{ponce2020sensor}. However, motion sensors still struggle with accuracy and with occlusions or furniture placed in a room. EEG headsets are also be extremely difficult to use in the real world.

The URFD dataset was recorded with two Kinect cameras. One was at eye level and one was ceiling-mounted for fall sequences \cite{kwolek2015improving}. However, only a single eye level camera was used to record the ADL activities. The accelerometer was worn on the lower back using an elastic belt. This sensor location is not ideal because a special device would need to be worn. The dataset contained only 30 falls and 40 activities of daily living. Along with the limited dataset size, it also contained short, segmented activities and falls with limited variation. 

The CMDFALL dataset \cite{Tran2018} focuses on multi-view as well as multi-modal capturing. Seven overlapping Kinect sensors were used, and two accelerometers were worn on the right hand and hip. The room was well lit, with a minimal amount of furniture in the room unless required for the fall scenario. It also contained eight falls and 12 actions. The eight falls were based on three main groups, walking, lying on the bed, or sitting on a chair. The dataset also mentioned many of the same issues such as trimmed videos, and limited fall styles. However, they included a limited variety of simple single action activities and fall styles. The environment also is always well lit and minimal furniture and occlusion in the space.

These existing multi-modal datasets lack consideration for lighting, furniture, fall, variety of ADL, and camera placement. This may impact real-world performance. Not only were these technical considerations lacking but also practical considerations with many multi-modal datasets requiring multiple wearable devices to be worn.

\begin{table*}[]
\caption{Highlighting various multi-modal datasets, their modalities and considerations}
\label{table:multimodaldatasets}
\centering
\resizebox{\linewidth}{!}{%
\begin{tabular}{|l|l|l|l|l|l|l|} 
\hline
\textbf{Datasets} & \begin{tabular}[c]{@{}l@{}}\textbf{Number of }\\\textbf{Modalities}\end{tabular} & \begin{tabular}[c]{@{}l@{}}\textbf{Number of }\\\textbf{Fall Types}\end{tabular} & \begin{tabular}[c]{@{}l@{}}\textbf{Number of }\\\textbf{Activities}\end{tabular} & \textbf{Participants} & \textbf{Trials} & \begin{tabular}[c]{@{}l@{}}\textbf{Length of}\\\textbf{~ADL (secs)}\end{tabular} \\ 
\hline
UR Fall
  Detection Dataset & 3 & 3 & 8 & 5 Adults & 70 & 10 \\ 
\hline
UP Fall
  Detection Dataset & 4 & 5 & 6 & 17 Adults & 70 & 10 - 60~ \\ 
\hline
CMDFALL & 3 & 8 & 12 & 50 Adults & 20 & 22.5 \\ 
\hline
\begin{tabular}[c]{@{}l@{}}Activity Recognition for \\Indoor Fall Detection Using\\Convolutional Neural Network\end{tabular} & 2 & 8 & 5 & 10 Adults & 20 & 30-60 \\ 
\hline
Our
  Dataset & 8 & NA & 25 & \begin{tabular}[c]{@{}l@{}}30 Adults\\10 Older Adults\end{tabular} & 400 & 180-240~ \\
\hline
\end{tabular}
}

\end{table*}


\subsection{Introduced Multi-modal Dataset (MUVIM)}

To circumvent the problems associated with previous uni-modal and multi-modal datasets, we present the Multi Visual Modality Fall Detection Dataset (MUVIM), which is a novel multi-camera multi-modality fall detection dataset. The MUVIM dataset places a larger emphasis on different types of camera based modalities. Multiple camera modalities were selected because of their practical implication compared to wearable or other types of modalities. Additionally, this allows for a direct comparison between privacy-protecting and non privacy protecting modalities. 

In MUVIM, wearable modalities were also included for completeness and to allow researchers to explore various combinations and performances. However, they were removed from further analysis.However, they were removed from further analysis in this paper. Camera modalities involving thermal, depth, infra-red and RGB cameras are only considered to develop models to detect falls.

The RGB camera is the most commonly available camera and provides a good baseline for comparison with the other modalities. Infra-red contains only grayscale images, but the image quality is still high enough to discern identifying features. However, these are low cost cameras, that operate well in low light scenarios with a wide field of view. Depth cameras are less common, but are still relatively low in cost. They are light-independent, and do not capture identifying features. They can struggle with creating a consistent image due to "holes" that occur in the depth-math due to sensor limitations. In addition, they struggle with scenarios in which participants, backgrounds or occlusions have similar distances. For access to the dataset, email: bing.ye@utoronto.ca. Please set the subject line too "Fall detection data access request", and include your include title, email address, work address, and affiliation. In addition, a data privacy waiver will have to be completed.

\section{Dataset Collection}

\subsection{Description of the Dataset}

\begin{figure}%
    \centering
    \includegraphics[width=0.45\textwidth]{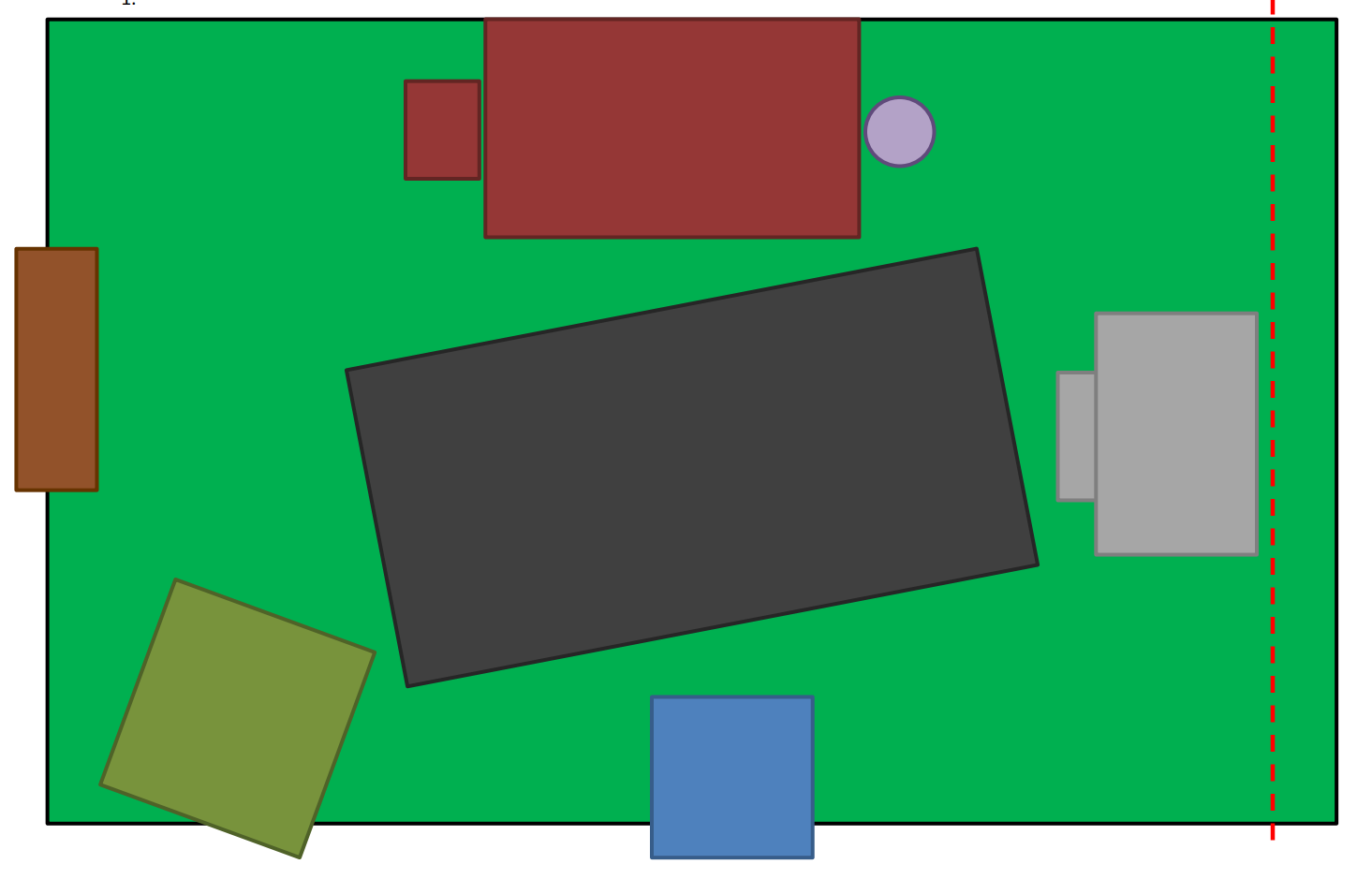}
    \caption{One example of the arrangements of five furniture pieces in room. Furniture pieces are as follows; \textbf{black} - mat; \textbf{brown} - shelf; \textbf{olive} - chair; \textbf{grey} - desk + chair; \textbf{red}blue - table; \textbf{red} - sofa + end table; \textbf{purple} - lamp. }%
    \label{fig:furniture_map}%
\end{figure}

Six (6) vision-based sensors were mounted on the ceiling of a floor at the Intelligent Assistive Technology and Systems Lab (IATSL), University of Toronto. This study was approved by the Research Ethics Board at the university. The participants have given their written consents to publish their images for research purposes. The layout of the lab is presented in Figure \ref{fig:furniture_map}. 

Four camera types were used:
\begin{enumerate}[label=(\roman*)]
\item One Hikvision IP network camera is a dome- style security camera that uses IR illumination. It captures 20 fps video at a resolution of 704 x 408 pixels. 
\item A StereoLabs ZED Depth Camera, captures 3D depth and RGB images. It releases on ambient illumination from the room to capture video at 30 fps and 1280 x 720 pixels. 
\item One Orbbec Astra Pro camera, captured 3D depth and IR images using near-IR illumination. It captured video at 30 fps and at 640 x 840 pixels. 
\item Three FLIR ONE Gen 3 Thermal cameras were used. These were used in order to capture an entire view of the room due to their narrow field of view. They work as an attachment to a smartphone and capture video at 8.7 fps at 1440 x 1080 pixels. 
\end{enumerate}

 All cameras were mounted in the middle of the ceiling except for the thermal cameras which had two more placed separately on each side of the room in order to cover the entire room. An Empatica wristband was worn by each participant to monitor movement and physiological signals. An overview of all modalities used in this study can be found in Table \ref{tab:devices}. Data was collected simultaneously from all sensors during each trial.

\begin{figure}
\centering
\begin{subfigure}[b]{.45\linewidth}
\includegraphics[width=\linewidth]{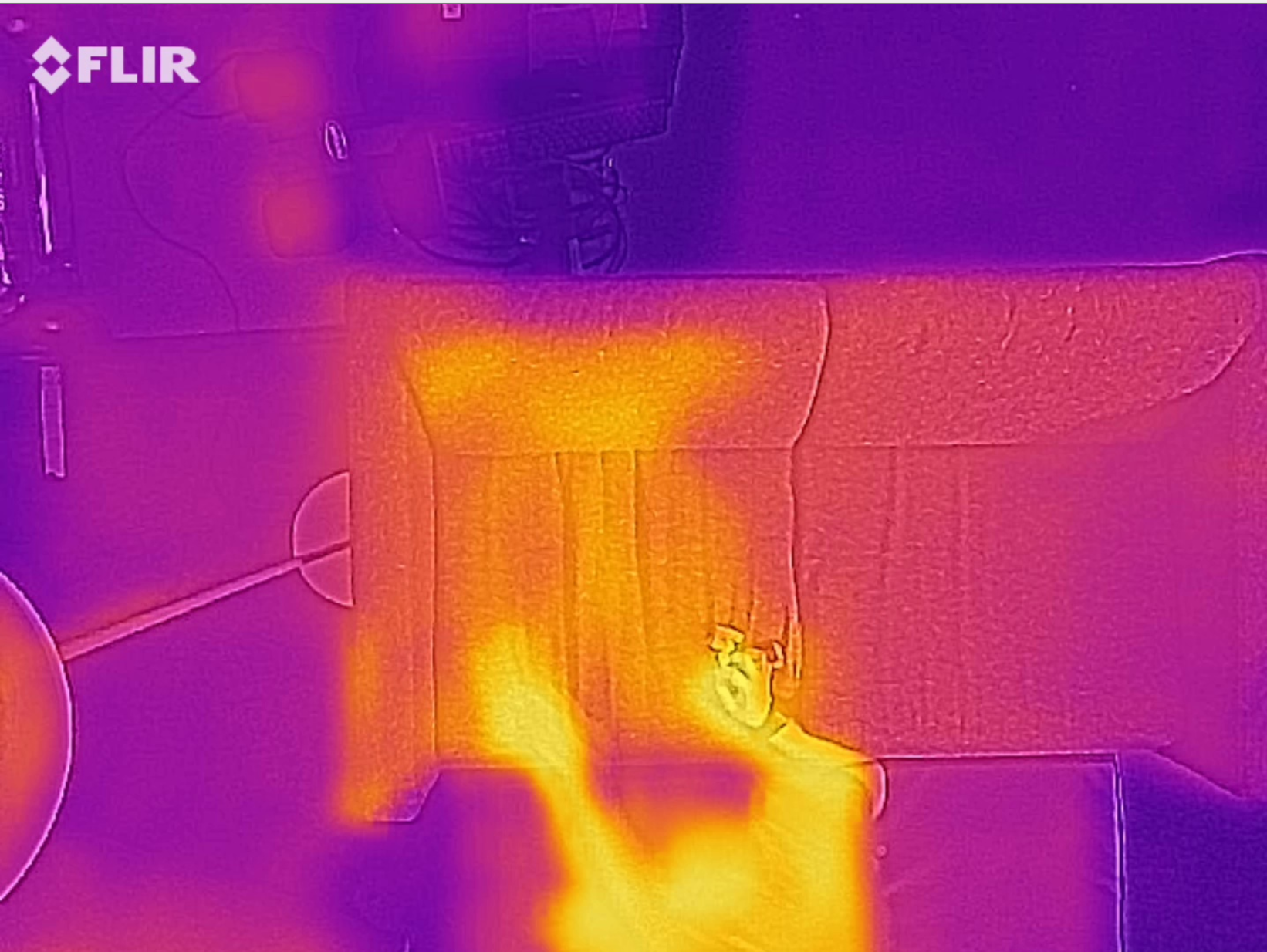}
\caption{Center FLIR Thermal}\label{fig:mouse}
\end{subfigure}
\begin{subfigure}[b]{.45\linewidth}
\includegraphics[width=\linewidth]{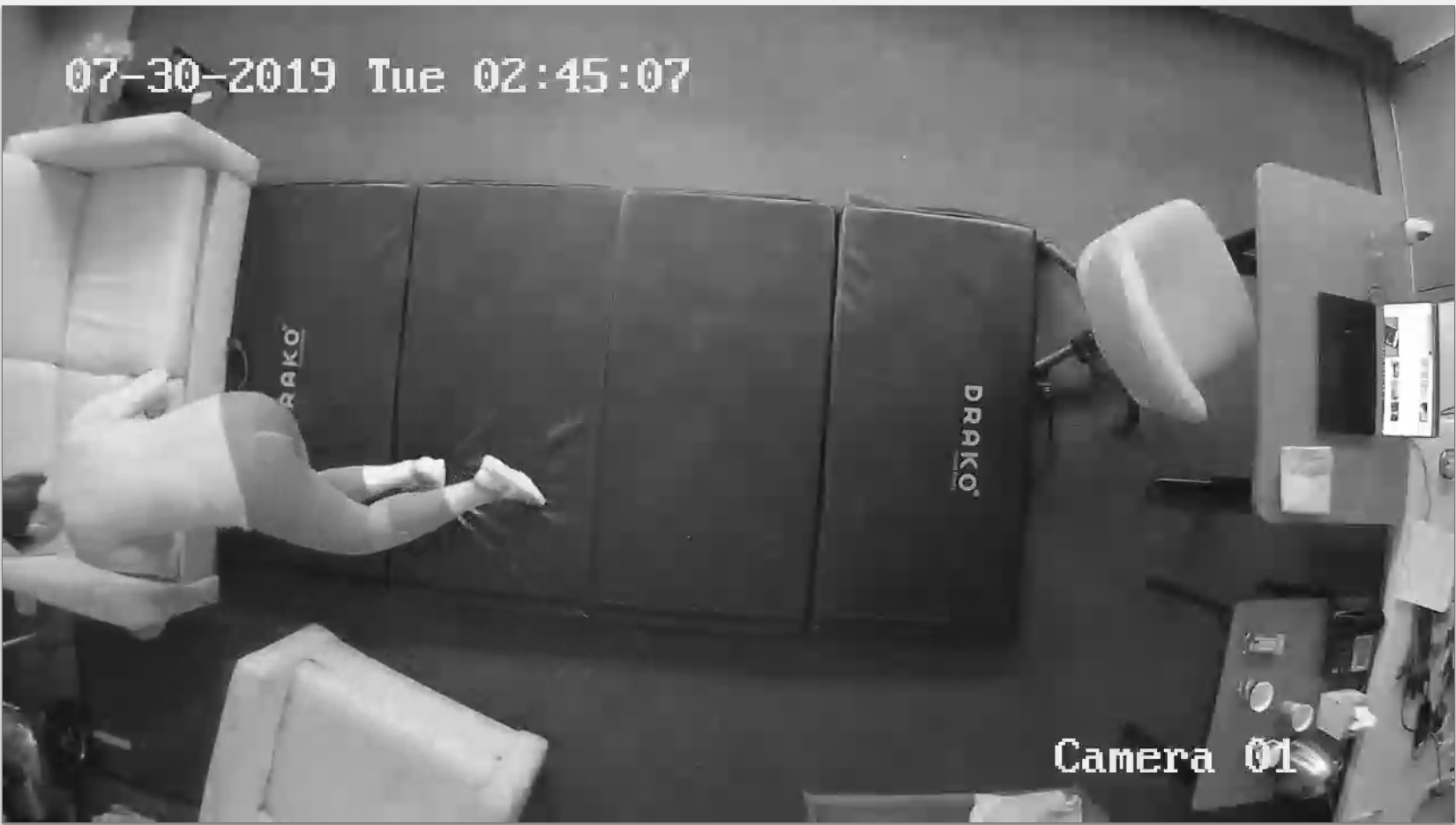}
\caption{Hikvision IP}\label{fig:mouse}
\end{subfigure}

\begin{subfigure}[b]{.45\linewidth}
\includegraphics[width=\linewidth]{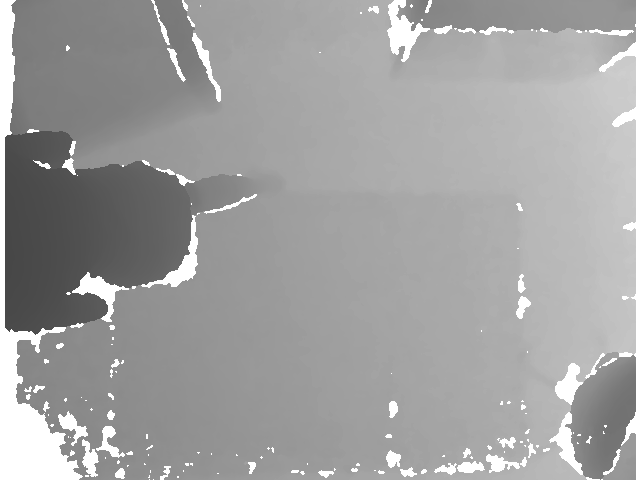}
\caption{Orbbec Depth}\label{fig:gull}
\end{subfigure}
\begin{subfigure}[b]{.45\linewidth}
\includegraphics[width=\linewidth]{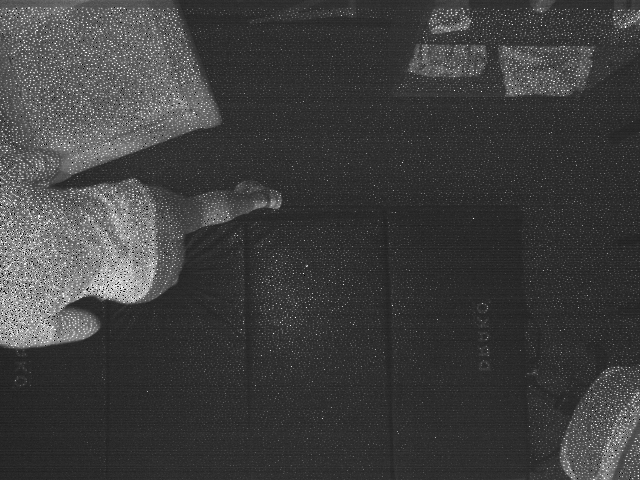}
\caption{Orbbec IR}\label{fig:tiger}
\end{subfigure}

\begin{subfigure}[b]{.45\linewidth}
\includegraphics[width=\linewidth]{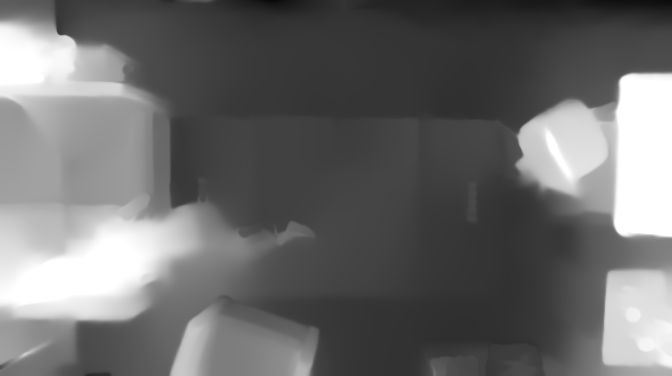}
\caption{Stereolabs ZED Depth}\label{fig:gull}
\end{subfigure}
\begin{subfigure}[b]{.45\linewidth}
\includegraphics[width=\linewidth]{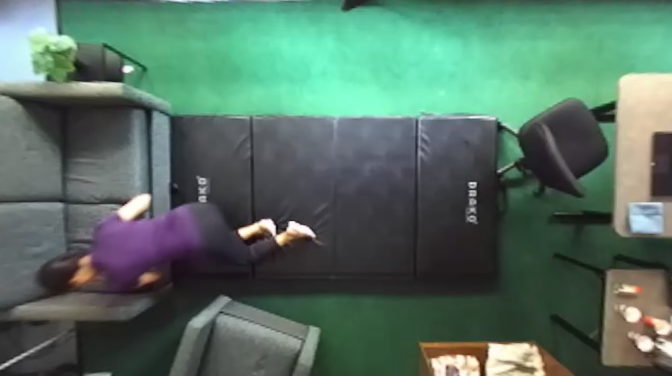}
\caption{Stereolabs ZED  RGB}\label{fig:tiger}
\end{subfigure}

\caption{Start of the fall as indicated by manually produced labels for each camera}
\label{fig:activity_2}
\end{figure}

\begin{table*}
\large
\centering
\caption{Highlighting devices used and the data collected form each.}
\label{tab:devices}
\resizebox{\textwidth}{!}{%
\begin{tabular}{|l|l|l|l|}
\hline
\textbf{Device}                                                                                                          & \textbf{Description}                                                                                      & \textbf{Installation}                                                    & \textbf{Data Collected}                                                                                                                 \\ \hline
Hikvision IP network camera (1)                                                                                          & \begin{tabular}[c]{@{}l@{}}Dome-style IP security camera \\ (IR illumination) \\ FOV: 106 degrees\end{tabular}                & Ceiling (centre)                                                         & \begin{tabular}[c]{@{}l@{}}IR video (grayscale)\\ File Format: mp4\\ Framerate (fps): 20\\ Resolution: 704 x 480 pixels  \end{tabular}    \\ \hline
StereoLabs ZED depth camera (1)                                                                                          & \begin{tabular}[c]{@{}l@{}}3D depth camera \\ (ambient illumination) \\ FOV: 90 x 60  degrees\end{tabular}                         & Ceiling (centre)                                                         & \begin{tabular}[c]{@{}l@{}}RGB video, Depth video\\ File Format: avi\\ Framerate (fps): 30\\ Resolution: 1280 x 720 pixels\end{tabular} \\ \hline
Orbbec Astra Pro (1)                                                                                                     & \begin{tabular}[c]{@{}l@{}}3D depth camera \\ (near-IR illumination) \\ FOV: 60 x 49.5 degrees\end{tabular}                         & Ceiling (centre)                                                         & \begin{tabular}[c]{@{}l@{}}IR video, Depth video\\ File Format: avi\\ Framerate (fps): 30\\ Resolution: 640 x 840\end{tabular}                    \\ \hline
\begin{tabular}[c]{@{}l@{}}FLIR ONE Gen 3 (3)\\ Model number: SM-G532M \\ (IATSL 278, IATSL 279, IATSL 280)\end{tabular} & \begin{tabular}[c]{@{}l@{}}Thermal camera attachment \\ for smartphones \\ FOV: 50 x 38 degrees\end{tabular}                      & \begin{tabular}[c]{@{}l@{}}Ceiling \\ (left, centre, right)\end{tabular} & \begin{tabular}[c]{@{}l@{}}Thermal video\\ File Format: mp4\\ Framerate: 8.7 fps\\ Resolution: 1440 x 1080 pixels\end{tabular}          \\ \hline
Empatica E4 (1)                                                                                                          & \begin{tabular}[c]{@{}l@{}}Wristband for monitoring of movement \\ and physiological signals\end{tabular} & Wearable                                                                 & \begin{tabular}[c]{@{}l@{}}Plain text data\\ File Format: csv\end{tabular}                                                              \\ \hline
\end{tabular}%
}

\end{table*}

\subsection{Participants and Activities}

The study was divided into two phases, each with a different participant population. In Phase 1, data was collected from 30 healthy younger adults between 18 and 30 years of age (mean age = 24; number of females = 14 (or 46.7\%). Inclusion criteria for this phase included: must be from ages 18-30, be clear of any health complications that may hinder balance or performance in the study, and must be able to understand and speak English, must be able to move around safely in a furnished room without eyeglasses, must be able to travel and attend sessions on site. In Phase 2, data was collected from 10 healthy older adults, who are at least 70 years of age (mean age =76.4 ; number of females = 5 (or 50\%). Participants in Phase 1 were asked to simulate falls onto a 4” thick crash mat. Participants in Phase 2 did not require the use of a crash mat, since no falls were simulated by this population.

Prior to a session, participants were provided with a consent form detailing the protocol of this study. Once consent was provided, the researchers confirmed the eligibility of the participant by conducting a brief screening questionnaire. Two versions of this questionnaire were used, since the populations for Phase I and Phase 2 were differed. In Phase 2, questions pertained to the participants mobility and vision issues to ensure safety and older adults were asked if they have experienced a fall within the last year to help ensure they do not have any mobility issues. 

Each session was approximately one (1) hour in length, including consent, preparation time, data collection, and intermissions to change the setting of the room. The setting of the room (including furniture and the crash mat) was randomized for each participant based on five (5) pre-made arrangements. The furniture and crash mat were stationary throughout each session, but were moved to a new configuration between participant sessions. This is done to avoid building a trivial classifier to detect crash mat as the cue for a fall in a scene. These five room settings were created in order to include as many props and furniture within the field of view of all modalities. 

Each session contained ten (10) trials, five "day-time" trials that were well lit and five "night-time" trials with poor lighting. Blackout curtains and main overhead lights were turned off in night time trials but an incandescent lamp was still left on in order to provide enough illumination to allow participants to move around safely. Each trial required the participant to act out a scripted “story”, while interacting with various furniture and props in the scene.For example, one story involves the participant returning home from an outing, putting down his or her bags, taking off his or her shoes and jacket, making tea, and sitting down at a computer. The format of each story was intended to put participants at ease and increase the realism of the data that was collected. The collection of stories used in this study covered a broad range of normal activities that might occur in a younger/older adult’s living room. In addition, the stories allowed the participants to move around the room as they interact with the furniture and props placed throughout.  The order of completion for all ten trials were randomized between participants to prevent researcher bias (ex. non-significant trends/presumptions due to order of trials) and allowed for better comparison between trials. 

\begin{figure}
    Five general representative fall types of older adults were selected 
    \begin{description}
      \item[$\bullet$ ] Tripping or stumbling forward, and falling when unable to catch oneself.
      \item[$\bullet$ ] Standing or walking, losing one's balance, and falling backwards or sideways. Note that a backward fall ends in a sitting or a lying position.
      \item[$\bullet$ ] Falling slowly/incrementally. (Person loses balance, catches self on some object, but continues to fall slowly when they don’t have the strength to recover.)
      \item[$\bullet$ ] Attempting to sit, but falling when the chair slides out from underneath.
      \item[$\bullet$ ] Attempting to stand, putting weight on a chair, table or walker, but the supporting object moves away and fails to provide support.
      \item[$\bullet$ ] Sitting in a chair, leaning forward to reach an item on the floor, putting on shoes, or attempting other activities, and toppling out of the chair.
    \end{description}
\end{figure}

At the beginning of each session, a research assistant (outside the view of the camera) would provide cue to the participant to perform certain activities. These cues were based on a pre-determined script that contained different scenarios (as discussed above). The participant would follow the cues (e.g. sit on sofa, walks around, pick an object from the floor, work on laptop, etc.) and perform these activities the way they would like to do. In each session, a person carries out normal activities based on the given cues and each session would end with a fall on the crash mat. The type of falls were kept diverse across each session and for different persons to capture different types of falls. Each story was assigned one or more fall types and recovery methods that could logically fit at the end of the story. If multiple fall types where possible for that story one was determined at random. A higher number of normal activities relative to fall events were deliberately performed to represent actual scenarios. A disproportionately high number of falls may over-simplify or bias the predictive models. In summary, a trial consisted of: enacting a story; simulating a fall; and engaging in a recovery, all of which were chosen randomly from a pre-defined script.

Participants in Phase 2 were also asked to complete ten stories but were not tasked with simulating a fall or a recovery. 

Participants were prompted through the steps of each trial. However, these prompts were not given with specific details to allow the participant to complete the assigned tasks based on their own interpretation. This would help in building generalized classifiers and not based on a particular sequence of activities.

The Empatica E4 wristband was turned on at the beginning of the session. Trials were defined through the pressing of the main button by the participant at the beginning and end of each trial. The Astra Pro, FLIR ONE, ZED, and IP cameras were manually controlled by the researchers. Recordings were started at the beginning of each trial, and stopped at the end of each trial.

\subsection{Data Pre-Processing and Consolidation}

\subsubsection{Data Structure}

\begin{figure}
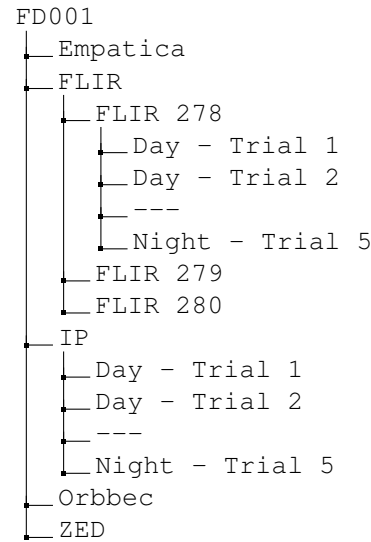

    \centering
    \begin{minipage}{7cm}
    \dirtree{%
    .1 FD001.
    .2 Empatica.
    .2 FLIR.
    .3 FLIR 278.
    .4 Day - Trial 1.
    .4 Day - Trial 2. 
    .4 ----- . 
    .4 Night - Trial 5. 
    .3 FLIR 279.
    .3 FLIR 280.
    .2 IP.
    .3 Day - Trial 1.
    .3 Day - Trial 2. 
    .3 ----- . 
    .3 Night - Trial 5.
    .2 Orbbec.
    .2 ZED.
    }
    \end{minipage}
    \caption{Directory Tree Used to Store Trials and Cameras. FD001 is the folder for participant 1, containing all cameras. Each camera has its own folder, containing 10 sub-folders for each trial done (labeled day or night). The FLIR thermal camera contains three additional folders for each three thermal cameras used (Note: 278, 279 and 280 are generated labels). }
    \label{fig:datatree}%
\end{figure}

Immediately after each session, the researchers transferred all collected data to a secure network server, and into a folder labelled with their participant ID. In each participant’s folder contained subfolders for each type of modality used during the trial including data from the Empatica E4 wristband. See Figure \ref{fig:datatree} for a  breakdown of the  directory structure of the public dataset. Each modality folder contained ten folders for each trial (ie. Day 1-5, Night 1-5). Once data transfer was complete, the researchers converted each of the camera videos into individual frames, in JPEG format.

\subsubsection{Labelling Procedure}

Once all data was transferred, the  researchers labelled the beginning and end frame numbers for each fall that occurred during all recorded sessions. All labels were noted using an Excel spreadsheet. Two researchers were recruited for this task to reduce sample bias and increase labelling accuracy. The start of a fall was marked as the frame when the participant started to lose balance. The end of a fall was marked as the frame when the participant was at rest on the ground.   The types of falls that were observed are listed below. Certain trials also contained more than one fall. In this case, both falls were documented in the spreadsheet.

\begin{figure*}[h!]
  \includegraphics[width=\textwidth]{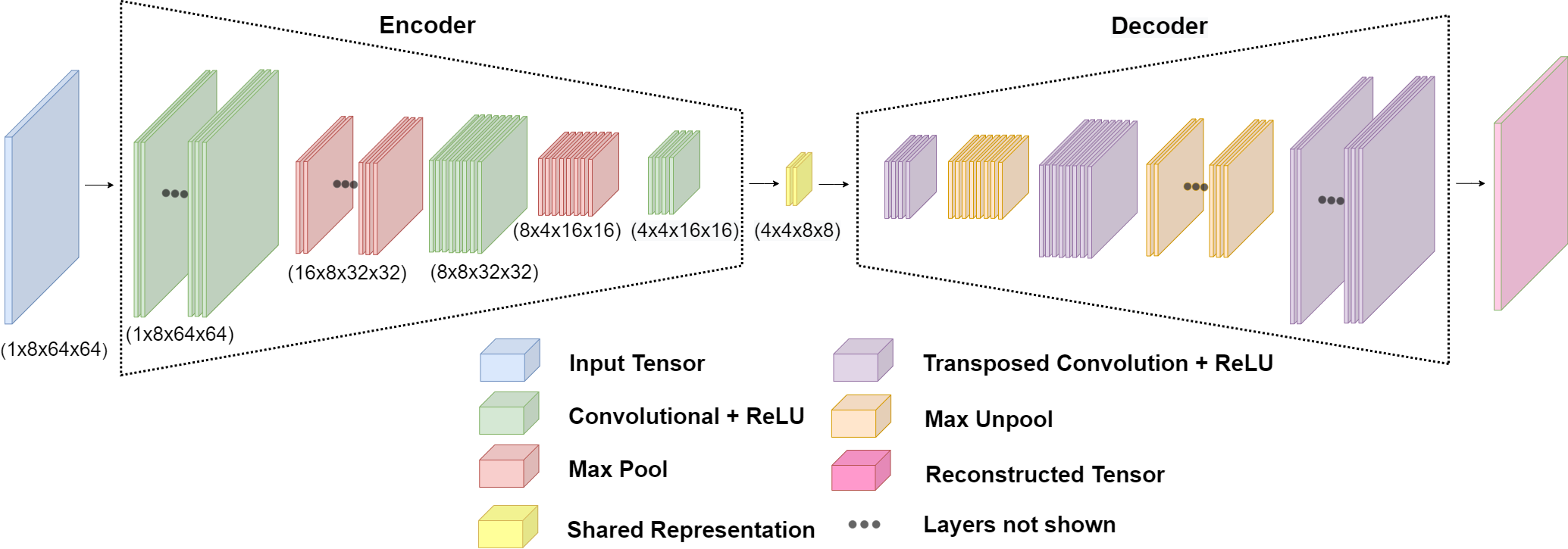}
  \caption{3D Autoencoder Structure}
  \label{fig:graphauto}%
\end{figure*}
 
\subsubsection{Limitations}

One major limitation involved the positioning of the FLIR cameras. The cameras were often not able to capture the complete duration of the fall clearly. For example, when the start of fall was unclear in the footage, the first visible frame was used instead. If the end of fall was clearly visible, it was still recorded in the main label table. This led to the incomplete labelling of falls for certain trials. Another limitation was that recording software for some modalities would froze during the trial, particularly with the thermal and Orbbec cameras, and thus were unable to capture the fall. However, this did not happen often and was not a major issue with all missing videos noted in the Appendix. For any trial that was missing the start or end of fall frames, the researcher made a note but did not record the frame into the data sheet. A further limitation appeared due to the lack of detail produced by recordings from the FLIR ONE cameras. Its low resolution video made it difficult to observe the exact start and end frames for certain falls. Specifically, we observed that when participants remained in one position for a long time, they left a heat residue, and thus made it difficult to differentiate between the residue and the participant who had fallen.

\section{Fall Detection}

To compare the performance of the camera, a baseline modality performance was required. This would allow to the comparison between camera modalities on the MUVIM fall detection dataset. 

Fall detection was approached as a one-class classification problem or anomaly detection owing to the rarity of fall events \cite{khan2014one}. Previous work has established that 3D CNN auto-encoders can achieve high performance by learning spatial and temporal features \cite{Nogas2020}. In this approach, a spatio-temporal autoencoder is trained only on video clips of normal activities that are available in abundance. During testing, the autoencoder should be able to reconstruct unseen normal activities with low reconstruction error. However, this autoencoder will reconstruct unseen falls with a higher reconstruction error. Thus, using 3D autoencoder, a fall can be effectively detected as an anomaly \cite{Nogas2020} We adapted the DeepFall \cite{Nogas2020} model and re-implemented it in PyTorch. The source code of the implementation is available at GitHub [link address].

A 3D CNN auto-encoder is formed through stacked layers of 3D convolutional and 3D max-pooling layers. Model parameters and design were adapted from previous work including, stride, padding, activation functions and kernel size \cite{Nogas2020}. Similar preprocessing steps were performed. All videos were resized to 64 x 64 pixels and interpolated or extrapolated to change the frame rate of each video to 8 fps. Interpolation was performed by duplicating existing frames. Depth videos were in-painted in order to fill out black sections of the video. These black sections occur from poor reflection of near-infrared light required to measure "time-of-flight" in order to determine depth. In-painting was done using an Navier-Stokes based in-paint function provided in the OpenCV python package. Model parameters were changed from the original DeepFall model to remove drop out layers from auto-encoders as they simply added noise to the encoder and did not provide any improved performance. Training was performed with 20 epochs as further training did not improve results. The MaxUnpool3d operation was used in PyTorch, which sets all non-maximal imputed values are set to zero. The architecture of the 3DCNN autoencoder models for fall detection is shown in Figure \ref{fig:graphauto}.However, batch size was increased to 128 frames and frame rate was lowered to eight fps. This decrease in frame rate efficiently increased the temporal window of the model. Experiments changing window size down to four did not significantly alter results. 

Training was performed on all videos of normal activities of daily living and then tested on videos that contain falls. A notable difference between the data splits is that the ADL videos only contained elderly participants, while the fall videos only contain younger adults. This results in 100 videos of older adults for training and 300 for testing of younger adults. However, due to data loss during capturing of data the number of videos for each modality varies. Since the thermal cameras have a narrow field of view, many falls were not captured within the central cameras field of view. When controlling for videos that are common across all modalities then only 182 falls for testing and 62 videos of ADL for training are available. Performance is reported for these videos. 
\begin{figure}
\centering
\label{fig:reconmethods}
\begin{subfigure}{.4\textwidth}
  \centering
  \includegraphics[width=\linewidth]{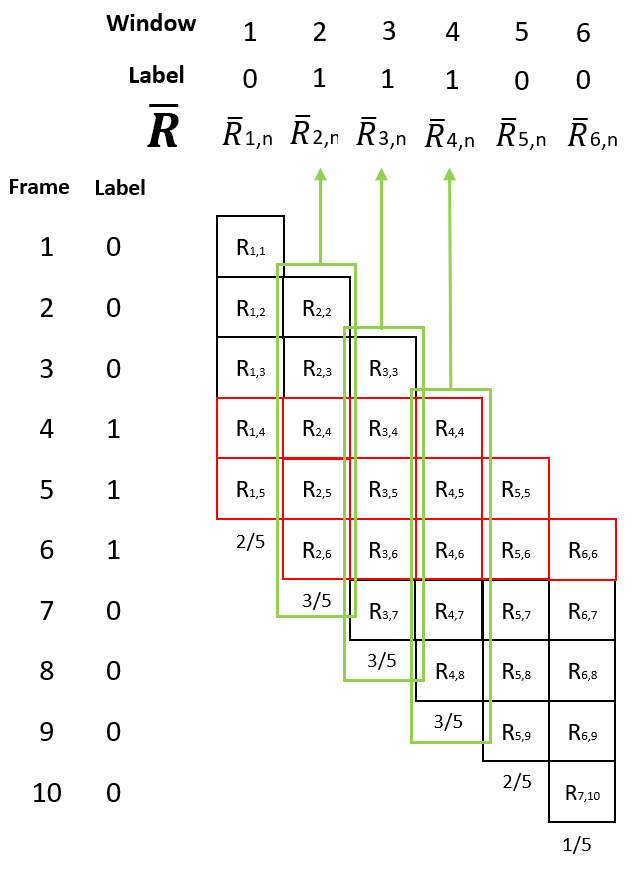}
  \caption{}
  \label{fig:sub1}
\end{subfigure}%

\begin{subfigure}{.4\textwidth}
  \centering
  \includegraphics[width=\linewidth]{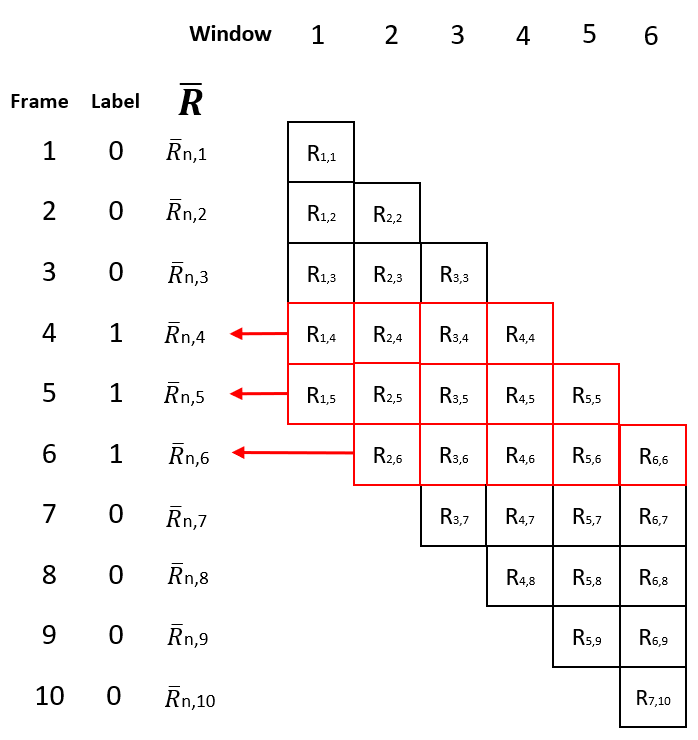}
  \caption{ }
  \label{fig:sub2}
\end{subfigure}
\caption{Within-Context versus Cross-Context scoring methods. \textbf{(A)} Within-context reconstruction scoring method. In this example, for illustration purposes three frames out of five must contain a fall for the window to be considered a fall. Reconstruction error is averaged across each window. \textbf{(B)} Cross-context reconstruction scoring method. Reconstruction error for a frame is averaged across all windows containing the frame. }
\end{figure}

As done in DeepFall, two main methods of determining reconstruction error can be used \cite{Nogas2020}. The first is on a per-frame basis (cross-context) and the second is on a per-window basis (within-context). These are outlined in Figure \ref{fig:reconmethods}. 

In cross-context, a reconstruction error is calculated for each individual frame. However, as a single frame is repeated in multiple windows, it has multiple reconstruction errors. The mean and standard deviation reconstruction error for a frame is then calculated across the multiple windows. As data is taken from across multiple windows it is called cross-context. AUC ROC is then calculated for each test video (per video) and then mean performance is reported (see Table \ref{table:mean_performannce_across_videos}). This allows for more specific tuning on a per video basis to a person and their specific activities. Alternatively, all reconsturction scores can be concatenated together, after which one AUC ROC can be found across all videos (global). Global threshold results are reported in Table \ref{table:global_cross_context}.This method may show a more generalizable level of performance. 

\begin{table}[htb]
\caption{Cross-context results when averaging performance on a per video basis. Where (\textsigma) is using the mean reconstruction error and (\textmu) is using the standard deviation of reconstruction error for each frame across all videos it appears in.}
\centering
\label{table:mean_performannce_across_videos}
\resizebox{\columnwidth}{!}{%
\begin{tabular}{!{}l!{\color{black}\vrule}l!{\color{black}\vrule}l!{\color{black}\vrule}l!{\color{black}\vrule}l!{}} 

Modalities & AUC ROC (\textsigma)      &  AUC ROC (\textmu)      &  AUC PR(\textsigma)      &  AUC PR(\textmu)        \\ 
\hline
IP         & \cellcolor{lightgray!75}0.950(0.049)  & 0.937(0.051) & 0.261 (0.280) & 0.199 (0.246)  \\ 
\hline
Orbbec IR     & 0.923 (0.085) &	\cellcolor{lightgray!75}0.927 (0.077) &	0.213 (0.224) &	0.195 (0.212)  \\ 
\hline
ZED Depth  & 0.874 (0.127) & \cellcolor{lightgray!75}0.894 (0.119) & 0.123 (0.145) & 0.143 (0.172) \\ 
\hline
Thermal  & 0.874 (0.112) &	\cellcolor{lightgray!75}0.883 (0.088) &	0.119 (0.11) &	0.124 (0.115)  \\ 
\hline
Orbbec Depth  & 0.839 (0.133) & \cellcolor{lightgray!75}0.872 (0.115) & 0.093 (0.138) & 0.101 (0.138) \\ 
\hline
ZED RGB    & \cellcolor{lightgray!75}0.859 (0.125) &	0.828 (0.123) &	0.082 (0.114) &	0.059 (0.097)  \\

\end{tabular}}
\end{table}

\begin{figure}%
    \centering
    \includegraphics[width=0.5\textwidth]{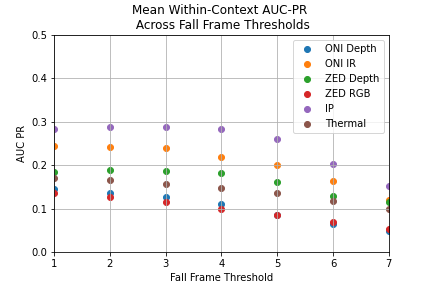}
    \caption{Mean Within-context AUC PR scores for various number of frames that contain a fall in order for the window to be considered to contain a fall.}%
    \label{fig:WithinContextGraphPR}%
\end{figure}

Secondly, a within-context anomaly score can be generated. This operates on a per-window basis, where mean reconstruction error of each sliding window is used. However, in order to determine a window's label a hyper-parameter on the number of falls must be set. For example, if a window contains eight frames, the hyper-parameter may be set to four, so that at least four frames of the eight must contain a fall for the window be classified as a fall. As such reporting for all within-context results is for all possible parameters given a window size of eight. The results are presented in Figure \ref{fig:WithinContextGraphPR}.

\begin{table}[htb]
\centering
\caption{Cross-context results when using a global receiving operating characteristic curve across all videos.}
\label{table:global_cross_context}
\resizebox{\columnwidth}{!}{%
\begin{tabular}{!{}l!{\color{black}\vrule}l!{\color{black}\vrule}l!{\color{black}\vrule}l!{\color{black}\vrule}l!{}} 

Modalities & AUC ROC (\textsigma)      & AUC ROC (\textmu)      & AUC PR (\textsigma)        & AUC PR (\textmu)        \\ 
\hline
IP         & 0.889 & \cellcolor{lightgray!75}0.913 & 0.076 & 0.067  \\ 
\hline
Orbbec IR     & 0.884  & \cellcolor{lightgray!75}0.902 & 0.058 & 0.066   \\ 
\hline
ZED Depth    & 0.780 & \cellcolor{lightgray!75}0.843 & 0.032 & 0.042  \\ 
\hline
Orbbec Depth  & 0.759 & \cellcolor{lightgray!75}0.831 & 0.022 & 0.031  \\ 
\hline
Thermal & 0.802 & \cellcolor{lightgray!75}0.814 & 0.050 & 0.056  \\ 
\hline
ZED RGB    & 0.619 & \cellcolor{lightgray!75}0.657 & 0.017 & 0.016  \\

\end{tabular}}
\end{table}

\begin{figure}%
    \centering
    \includegraphics[width=0.5\textwidth]{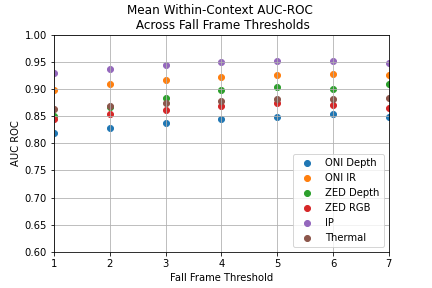}
    \caption{Mean Within-context AUC ROC scores for various number of frames that contain a fall in order for the window to be considered to contain a fall.}%
    \label{fig:WithinContextGraphROC}%
\end{figure}

 \begin{table}[]
\centering
\caption{Cross-context precision-recall results for a global threshold across all videos}
\label{table:withincontextbaselinepr}
\resizebox{\columnwidth}{!}{%
\begin{tabular}{!{}l!{\color{black}\vrule}l!{\color{black}\vrule}l!{\color{black}\vrule}l!{\color{black}\vrule}l!{\color{black}\vrule}l!{\color{black}\vrule}l!{\color{black}\vrule}l!{}} 
Fall Threshold & 1     & 2     & 3     & 4     & 5     & 6     & 7     \\
\hline
Baseline        & 0.018 & 0.016 & 0.013 & 0.011 & 0.008 & 0.006 & 0.004  \\
\hline
IP              & 0.282 & 0.288 & 0.288 & 0.283 & 0.259 & 0.202 & 0.153  \\
\hline
Orbbec IR          & 0.243 & 0.242 & 0.239 & 0.219 & 0.200 & 0.163 & 0.119 \\
\hline
ZED Depth       & 0.185 & 0.188 & 0.187 & 0.181 & 0.161 & 0.130 & 0.114  \\
\hline
Thermal         & 0.171 & 0.165 & 0.157 & 0.147 & 0.135 & 0.117 & 0.099  \\
\hline
ZED RGB         & 0.149 & 0.146 & 0.144 & 0.137 & 0.128 & 0.107 & 0.088  \\
\hline
Orbbec Depth       & 0.145 & 0.136 & 0.125 & 0.111 & 0.086 & 0.064 & 0.048 
\end{tabular}}
\end{table}

\section{Results and Discussion}

We observe good performance across all modalities, mean performance for each modality cross-contexts is reported in Table \ref{table:mean_performannce_across_videos}. The best performance is seen by the IP infra-red camera, achieving an average AUC ROC(\textsigma) of 0.95 (0.049) using cross-context mean reconstruction error. Performance was comparable between mean reconstruction error and standard deviation of reconstruction error, however most modalities performed slightly better using the standard error of reconstruction. Precision-recall performance is relative to the ratio of true positives in the dataset divided by the number of samples. Our dataset contained roughly 885 falls frames from a total of 92,709 frames for each modality. This means that the baseline of a random classifier would achieve a AUC PR of 0.0096. Best AUC PR performance is seen with IP modality achieving a AUC PR(\textsigma) of 0.261(0.280) standard deviation. 

Instead of averaging performance across all videos, reconstruction error from each video camera is concatenated into a single vector; AUC ROC and PR can then calculated. Table \ref{table:global_cross_context} presents the results. We observe a decrease in performance across all modalities. However, the strongest performing modalities had the least decline, with the IP camera sill achieving an AUC ROC (\textsigma) of 0.913 (down from 0.937(0.051). However, weaker modalities saw a larger decrease, specifically with the RGB camera decreasing from an AUC ROC (\textsigma) of 0.828(0.123)  to 0.657. When using global thresholding, a large decrease in the AUC PR performance is also seen. The best AUC PR performance decreased from the IP camera's AUC PR (\textmu) of 0.261(0.280) to 0.076. However, since the baseline for AUC PR is 0.0096 (due to the large class imbalance) performance is still better than a random classifier. This decrease in AUC PR is most likely caused by an optimal threshold for reconstruction error that cannot be found on a per video, but instead could be generalizable across all videos.

The within-context results using mean reconstruction error are shown in Figure \ref{fig:WithinContextGraphROC}. Similar performance is achieved with an within-context context scoring method. Performance increases as the number of falls in a window required to classify it a fall window increases, peaking at five and then beginning to slightly decrease. However, the precision-recall performance is best at a lower number. As a larger number of frames are required to classify a window as containing a fall, the fewer number of windows are considered to have a fall. Thus, the ratio of true positive labels to total labels will decrease. As baseline performance is too small to show clearly on the graph it is reported in Table \ref{table:withincontextbaselinepr}.

IR is the best performing modality (see Table \ref{table:mean_performannce_across_videos} with the IP camera achieving AUC ROC (\textsigma) scores of 0.950(0.049) AUC ROC (\textmu) scores of 0.937(0.077) and the Orbbec camera achieving AUC ROC (\textsigma) 0.923(0.085) and AUC ROC (\textmu) 0.927(0.077). IR cameras benefit from high visual clarity while also being able perform well in a range of lighting conditions. This creates a low amount of a noise within the image and less challenging conditions for an autoencoder to reconstruct. These factors can help improve the performance of the classifier. We also see a negligible difference between the IP camera and Orbbec IR camera’s indicating that achieved performance is due to modality differences rather camera variables such as recording frame rate or field of view. 

Thermal video also performed very well achieving an AUC ROC (\textsigma) of 0.874(0.112) and an AUC ROC (\textmu) of 0.883(0.088)  (see Table \ref{table:mean_performannce_across_videos}). This was despite the lower frame and narrower frame of view. This modality tends to obscure facial features, details in the background and create a silhouette of participants. This creates a strong separation between an individual and their surroundings. However, when an individual leaves or enters the frame the gradient in the image changes to represent the hottest and coldest parts of an image. As seen between images in Figure \ref{fig:activity_1} without a participant and Figure \ref{fig:activity_2} with a participant. This results may have resulted in a decrease performance due to increased variability between scenes and as a participants enter and leave frames. Additionally, this modality did include interpolated frames in order to increase FPS from 4 to 8.  

The ZED Depth camera achieved AUC ROC(\textmu) 0.894(0.119). The Orbbec Depth camera achieved 0.839 (0.133) and an AUC ROC(\textmu) of 0.872(0.115). Performance decreased for the depth modality with the use of a global threshold as compared to a per-video basis with an AUC ROC(\textmu) of 0.843 and 0.831 for the ZED and Orbbec cameras respectively. This may be due to depth cameras often have depth errors at an objects edges/corners, requiring in-painting. Despite in-painting improving performance, it is an estimation of missing values and thus introduces additional noise into the image. 

The RGB camera had the worst performance amongst all datasets with an AUC ROC (\textsigma) of  0.859(0.125) and an AUC ROC (\textmu) of 0.828(0.123) . The field of view, camera placement and image quality were similar to those of other datasets. Varied lighting conditions both within a single image and across videos may have greatly affected performance. 

As we approached one-class classification through measuring reconstruction error, having a lower amount of noise can help isolate reconstruction error due to fall activities. However, strong performance is still observed across various modalities and camera types, indicating that the signal strength is strong.

\section{Conclusions and Future Work}

In this paper, we present a multi-modal fall detection dataset with real-world considerations. It contains six-vision based modalities, and four physiological modalities; considering environmental factors, a variety of complex activities and privacy considerations. Performance across multiple visual modalities was analyzed within an anomaly detection framework. 

Infra-red modalities outperformed other modalities, followed by thermal and, then depth with traditional RGB, performing the worst. This order of results was maintained through various different reconstruction scoring methods (cross-context and within-context). This difference became more apparent when a global threshold was used to classify the results. It is also encouraging that strong privacy performing modalities such as the thermal and depth had competitive performance. This provides a path forward to create a strong performing and privacy protecting passive fall detection systems. 

Future work would expand on this analysis through the use of multi-modal fusion. By being able to combine multiple modalities and their respective strengths it may improve overall performance with reduced false positives. Additionally, the objective or loss functions of the autoencoder may be altered in order to improve the reconstructive ability of the autoencoder. Future work may also explore other deep learning methods, such as applying contrastive learning or attention through the use of visual transformers. 

We believe MUVIM will help provide a new bench mark dataset and help drive the development of real world fall detection systems that can be deployed effectively in peoples homes.  

\section{CONFLICT OF INTEREST STATEMENT}
The authors declare that the research was conducted in the absence of any commercial or financial relationships that could be construed as a potential conflict of interest.

\section{ACKNOWLEDGMENTS}
The authors would like to thank all participants for their time, and efforts in the creation of this dataset. We would also like to thank Paris Roserie for his support in collection of the dataset.

\pagebreak 
\pagebreak 
\bibliography{main.bib}

\begin{thebibliography}{10}
\providecommand{\url}[1]{#1}
\csname url@samestyle\endcsname
\providecommand{\newblock}{\relax}
\providecommand{\bibinfo}[2]{#2}
\providecommand{\BIBentrySTDinterwordspacing}{\spaceskip=0pt\relax}
\providecommand{\BIBentryALTinterwordstretchfactor}{4}
\providecommand{\BIBentryALTinterwordspacing}{\spaceskip=\fontdimen2\font plus
\BIBentryALTinterwordstretchfactor\fontdimen3\font minus
  \fontdimen4\font\relax}
\providecommand{\BIBforeignlanguage}[2]{{%
\expandafter\ifx\csname l@#1\endcsname\relax
\typeout{** WARNING: IEEEtran.bst: No hyphenation pattern has been}%
\typeout{** loaded for the language `#1'. Using the pattern for}%
\typeout{** the default language instead.}%
\else
\language=\csname l@#1\endcsname
\fi
#2}}
\providecommand{\BIBdecl}{\relax}
\BIBdecl

\bibitem{WHOFalls2021}
\BIBentryALTinterwordspacing
``{Falls},'' April 2021. [Online]. Available:
  \url{https://www.who.int/news-room/fact-sheets/detail/falls}
\BIBentrySTDinterwordspacing

\bibitem{KramarowEChenLHHedegaardH2015}
H.~H. Warner M. Kramarow~E, Chen~LH, ``Deaths from unintentional injury among
  adults aged 65 and over: United states, 2000-2013.'' \emph{NCHS Data Brief},
  p. 199, 2015.

\bibitem{centers2006fatalities}
C.~for Disease~Control, Prevention \emph{et~al.}, ``Fatalities and injuries
  from falls among older adults---united states, 1993--2003 and 2001--2005,''
  \emph{MMWR: Morbidity and mortality weekly report}, vol.~55, no.~45, pp.
  1221--1224, 2006.

\bibitem{gillespie2012interventions}
L.~D. Gillespie, M.~C. Robertson, W.~J. Gillespie, C.~Sherrington, S.~Gates,
  L.~Clemson, and S.~E. Lamb, ``Interventions for preventing falls in older
  people living in the community,'' \emph{Cochrane database of systematic
  reviews}, no.~9, 2012.

\bibitem{Stinchcombe2014}
A.~Stinchcombe, N.~Kuran, and S.~Powell, ``{Seniors' falls in Canada: Second
  report: Key highlights},'' Tech. Rep. 2-3, 2014.

\bibitem{Rubenstein2002}
L.~Z. Rubenstein and K.~R. Josephson, ``The epidemiology of falls and
  syncope,'' \emph{Clinics in geriatric medicine}, vol.~18, no.~2, pp.
  141--158, 2002.

\bibitem{Tinetti1993}
M.~E. Tinetti, W.-L. Liu, and E.~B. Claus, ``Predictors and prognosis of
  inability to get up after falls among elderly persons,'' \emph{Jama}, vol.
  269, no.~1, pp. 65--70, 1993.

\bibitem{Igual2013}
R.~Igual, C.~Medrano, and I.~Plaza, ``Challenges, issues and trends in fall
  detection systems,'' \emph{Biomedical engineering online}, vol.~12, no.~1,
  pp. 1--24, 2013.

\bibitem{Khan2017}
S.~S. Khan and J.~Hoey, ``Review of fall detection techniques: A data
  availability perspective,'' \emph{Medical engineering \& physics}, vol.~39,
  pp. 12--22, 2017.

\bibitem{stone2014fall}
E.~E. Stone and M.~Skubic, ``Fall detection in homes of older adults using the
  microsoft kinect,'' \emph{IEEE journal of biomedical and health informatics},
  vol.~19, no.~1, pp. 290--301, 2014.

\bibitem{debard2012camera}
G.~Debard, P.~Karsmakers, M.~Deschodt, E.~Vlaeyen, E.~Dejaeger, K.~Milisen,
  T.~Goedem{\'e}, B.~Vanrumste, and T.~Tuytelaars, ``Camera-based fall
  detection on real world data,'' in \emph{Outdoor and large-scale real-world
  scene analysis}.\hskip 1em plus 0.5em minus 0.4em\relax Springer, 2012, pp.
  356--375.

\bibitem{Khan2016}
S.~Khan, ``{Classification and Decision-Theoretic Framework for Detecting and
  Reporting Unseen Falls. Thesis. University of Waterloo. Ontario, Canada.}''
  Tech. Rep., 2016.

\bibitem{charfi2012definition}
I.~Charfi, J.~Miteran, J.~Dubois, M.~Atri, and R.~Tourki, ``Definition and
  performance evaluation of a robust svm based fall detection solution,'' in
  \emph{2012 eighth international conference on signal image technology and
  internet based systems}.\hskip 1em plus 0.5em minus 0.4em\relax IEEE, 2012,
  pp. 218--224.

\bibitem{Tran2018}
T.-H. Tran, T.-L. Le, D.-T. Pham, V.-N. Hoang, V.-M. Khong, Q.-T. Tran, T.-S.
  Nguyen, and C.~Pham, ``A multi-modal multi-view dataset for human fall
  analysis and preliminary investigation on modality,'' in \emph{2018 24th
  International Conference on Pattern Recognition (ICPR)}.\hskip 1em plus 0.5em
  minus 0.4em\relax IEEE, 2018, pp. 1947--1952.

\bibitem{Chaudhuri2014}
S.~Chaudhuri, H.~Thompson, and G.~Demiris, ``Fall detection devices and their
  use with older adults: a systematic review,'' \emph{Journal of geriatric
  physical therapy (2001)}, vol.~37, no.~4, p. 178, 2014.

\bibitem{Chaudhuri2015}
S.~Chaudhuri, D.~Oudejans, H.~J. Thompson, and G.~Demiris, ``Real world
  accuracy and use of a wearable fall detection device by older adults,''
  \emph{Journal of the American Geriatrics Society}, vol.~63, no.~11, p. 2415,
  2015.

\bibitem{Mubashir2013}
M.~Mubashir, L.~Shao, and L.~Seed, ``{A survey on fall detection: Principles
  and approaches},'' \emph{Neurocomputing}, vol. 100, pp. 144--152, jan 2013.

\bibitem{Wang2020}
X.~Wang, J.~Ellul, and G.~Azzopardi, ``Elderly fall detection systems: A
  literature survey,'' \emph{Frontiers in Robotics and AI}, vol.~7, p.~71,
  2020.

\bibitem{Martinez-Villasenor2019}
L.~Mart{\'{i}}nez-Villase{\~{n}}or, H.~Ponce, J.~Brieva, E.~Moya-Albor,
  J.~N{\'{u}}{\~{n}}ez-Mart{\'{i}}nez, and C.~Pe{\~{n}}afort-Asturiano,
  ``{Up-fall detection dataset: A multimodal approach},'' \emph{Sensors
  (Switzerland)}, vol.~19, no.~9, 2019.

\bibitem{Pierleoni2016}
P.~Pierleoni, A.~Belli, L.~Maurizi, L.~Palma, L.~Pernini, M.~Paniccia, and
  S.~Valenti, ``{A Wearable Fall Detector for Elderly People Based on AHRS and
  Barometric Sensor},'' pp. 6733--6744, 2016.

\bibitem{Lee2005}
T.~Lee and A.~Mihailidis, ``{An intelligent emergency response system:
  Preliminary development and testing of automated fall detection},''
  \emph{Journal of Telemedicine and Telecare}, vol.~11, no.~4, pp. 194--198,
  2005.

\bibitem{Chaudhuri2017}
S.~Chaudhuri, L.~Kneale, T.~Le, E.~Phelan, D.~Rosenberg, H.~Thompson, and
  G.~Demiris, ``{Older Adults' Perceptions of Fall Detection Devices},''
  \emph{Journal of Applied Gerontology}, vol.~36, no.~8, pp. 915--930, aug
  2017.

\bibitem{Ma2014}
X.~Ma, H.~Wang, B.~Xue, M.~Zhou, B.~Ji, and Y.~Li, ``Depth-based human fall
  detection via shape features and improved extreme learning machine,''
  \emph{IEEE journal of biomedical and health informatics}, vol.~18, no.~6, pp.
  1915--1922, 2014.

\bibitem{Riquelme}
F.~Riquelme, C.~Espinoza, T.~Rodenas, J.-G. Minonzio, and C.~Taramasco,
  ``ehomeseniors dataset: An infrared thermal sensor dataset for automatic fall
  detection research,'' \emph{Sensors}, vol.~19, no.~20, p. 4565, 2019.

\bibitem{gutierrez2021comprehensive}
J.~Guti{\'e}rrez, V.~Rodr{\'\i}guez, and S.~Martin, ``Comprehensive review of
  vision-based fall detection systems,'' \emph{Sensors}, vol.~21, no.~3, p.
  947, 2021.

\bibitem{ramachandran2020survey}
A.~Ramachandran and A.~Karuppiah, ``A survey on recent advances in wearable
  fall detection systems,'' \emph{BioMed research international}, vol. 2020,
  2020.

\bibitem{auvinet2010fall}
E.~Auvinet, F.~Multon, A.~Saint-Arnaud, J.~Rousseau, and J.~Meunier, ``Fall
  detection with multiple cameras: An occlusion-resistant method based on 3-d
  silhouette vertical distribution,'' \emph{IEEE transactions on information
  technology in biomedicine}, vol.~15, no.~2, pp. 290--300, 2010.

\bibitem{Zhang2014}
Z.~Zhang, C.~Conly, and V.~Athitsos, ``Evaluating depth-based computer vision
  methods for fall detection under occlusions,'' in \emph{International
  Symposium on Visual Computing}.\hskip 1em plus 0.5em minus 0.4em\relax
  Springer, 2014, pp. 196--207.

\bibitem{Baldewijns2016}
G.~Baldewijns, G.~Debard, G.~Mertes, B.~Vanrumste, and T.~Croonenborghs,
  ``Bridging the gap between real-life data and simulated data by providing a
  highly realistic fall dataset for evaluating camera-based fall detection
  algorithms,'' \emph{Healthcare technology letters}, vol.~3, no.~1, pp. 6--11,
  2016.

\bibitem{vadivelu2016thermal}
S.~Vadivelu, S.~Ganesan, O.~R. Murthy, and A.~Dhall, ``Thermal imaging based
  elderly fall detection,'' in \emph{Asian Conference on Computer
  Vision}.\hskip 1em plus 0.5em minus 0.4em\relax Springer, 2016, pp. 541--553.

\bibitem{luo2012design}
X.~Luo, T.~Liu, J.~Liu, X.~Guo, and G.~Wang, ``Design and implementation of a
  distributed fall detection system based on wireless sensor networks,''
  \emph{EURASIP Journal on Wireless Communications and Networking}, vol. 2012,
  no.~1, pp. 1--13, 2012.

\bibitem{Gasparrini2014}
S.~Gasparrini, E.~Cippitelli, S.~Spinsante, and E.~Gambi, ``{A depth-based fall
  detection system using a Kinect{\textregistered} sensor},'' \emph{Sensors
  (Switzerland)}, vol.~14, no.~2, pp. 2756--2775, 2014.

\bibitem{martinez2020design}
L.~Martinez-Villase{\~n}or and H.~Ponce, ``Design and analysis for fall
  detection system simplification,'' \emph{JoVE (Journal of Visualized
  Experiments)}, no. 158, p. e60361, 2020.

\bibitem{ponce2020sensor}
H.~Ponce, L.~Mart{\'\i}nez-Villase{\~n}or, and J.~Nu{\~n}ez-Mart{\'\i}nez,
  ``Sensor location analysis and minimal deployment for fall detection
  system,'' \emph{IEEE Access}, vol.~8, pp. 166\,678--166\,691, 2020.

\bibitem{kwolek2015improving}
B.~Kwolek and M.~Kepski, ``Improving fall detection by the use of depth sensor
  and accelerometer,'' \emph{Neurocomputing}, vol. 168, pp. 637--645, 2015.

\bibitem{khan2014one}
S.~S. Khan and M.~G. Madden, ``One-class classification: taxonomy of study and
  review of techniques,'' \emph{The Knowledge Engineering Review}, vol.~29,
  no.~3, pp. 345--374, 2014.

\bibitem{Nogas2020}
\BIBentryALTinterwordspacing
J.~Nogas, S.~S. Khan, and A.~Mihailidis, ``{DeepFall: Non-Invasive Fall
  Detection with Deep Spatio-Temporal Convolutional Autoencoders},''
  \emph{Journal of Healthcare Informatics Research}, vol.~4, no.~1, pp. 50--70,
  mar 2020. [Online]. Available:
  \url{https://link.springer.com/article/10.1007/s41666-019-00061-4}
\BIBentrySTDinterwordspacing

\end{thebibliography}

\pagebreak 
\section{Appendix}

Participants with incomplete data: 9/30
\begin{itemize}[noitemsep]
    \item FD003:
    \begin{itemize}[noitemsep]
        \item Orbecc: Day 2, 4 (video stopped recording prematurely, doesn’t contain fall)
        \item FLIR 279: Day 2, 3 (video missing from drive, not transferred properly after trial)
        \item FLIR 280: Day 1,2,3 (video missing from drive, not transferred properly after trial)
    \end{itemize}
    \item FD006:
    \begin{itemize}[noitemsep]
        \item Orbecc: Day 3 (video missing from drive, not transferred properly after trial)
    \end{itemize}
    \item FD007:
    \begin{itemize}[noitemsep]
        \item Orbecc: Day 4 (video missing from drive, not transferred properly after trial)
    \end{itemize}
    \item FD009:
    \begin{itemize}[noitemsep]
        \item FLIR 280: Day 1 (video is corrupted on drive)
    \end{itemize}
    \item FD010:
    \begin{itemize}[noitemsep]
        \item Orbecc: Night 1,2,3,4,5 (invalid video on drive <ie. very small video size>, not transferred properly after trial)
    \end{itemize}
    
    \item FD029:
    \begin{itemize}[noitemsep]
        \item FLIR 279: ALL (video missing from drive, not transferred properly after trial)
    \end{itemize}
    
    \item FD030:
    \begin{itemize}[noitemsep]
        \item FLIR 279: ALL (Wrong participant recorded in video, not transferred properly after trial)
    \end{itemize}
    
    \item FD031:
    \begin{itemize}[noitemsep]
        \item Day 1,2,3,4,5, Night 1,2,3 (invalid video on drive <ie. very small video size>, not transferred properly after trial)
    \end{itemize}

    \item FD033:
    \begin{itemize}[noitemsep]
        \item Orbecc: Day 3,4,5 (invalid video on drive <ie. very small video size>, not transferred properly after trial)
    \end{itemize}

\end{itemize}

\begin{figure}
\centering
\begin{subfigure}[b]{\linewidth}
\includegraphics[width=\linewidth]{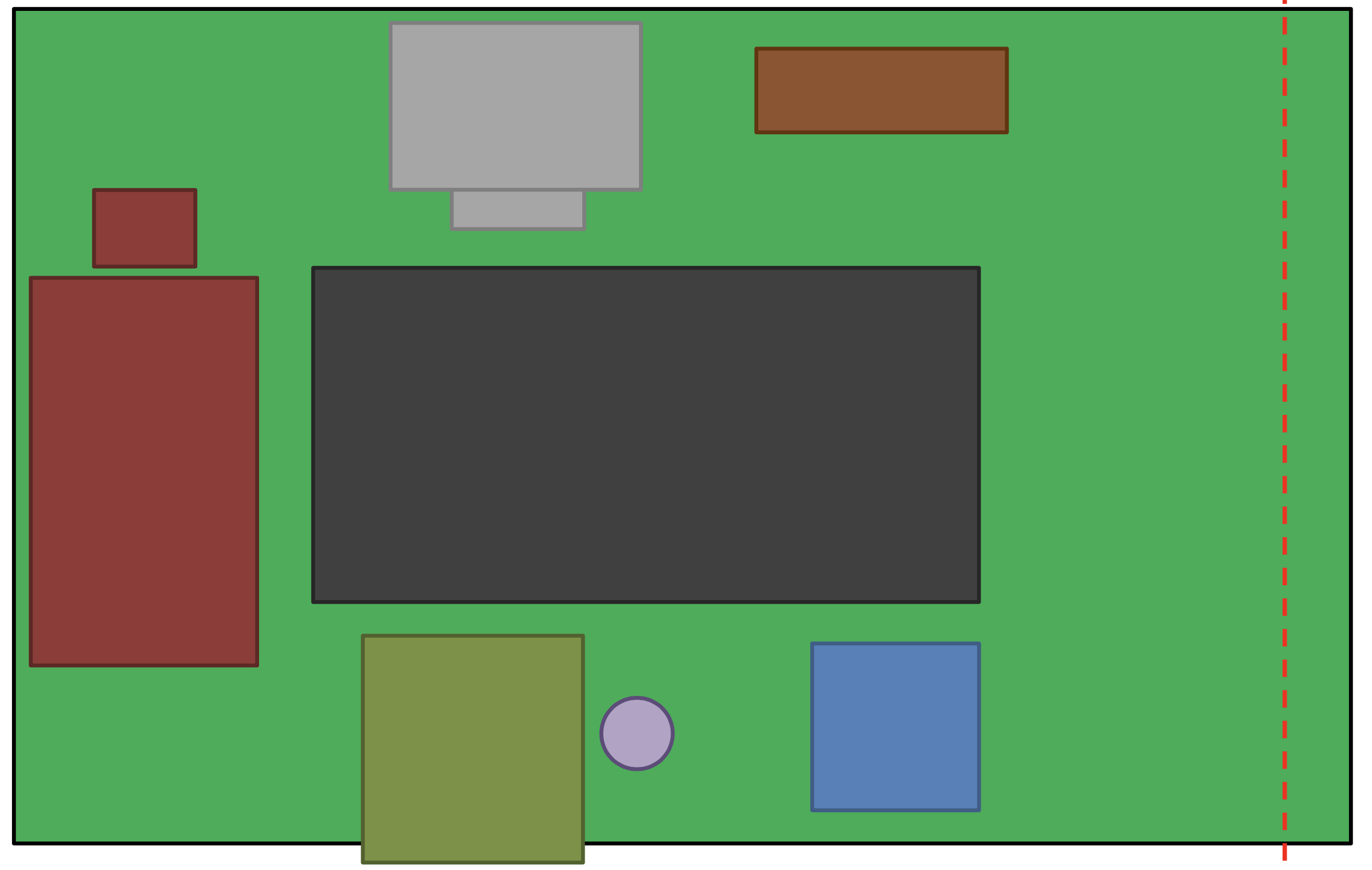}
\end{subfigure}

\begin{subfigure}[b]{\linewidth}
\includegraphics[width=\linewidth]{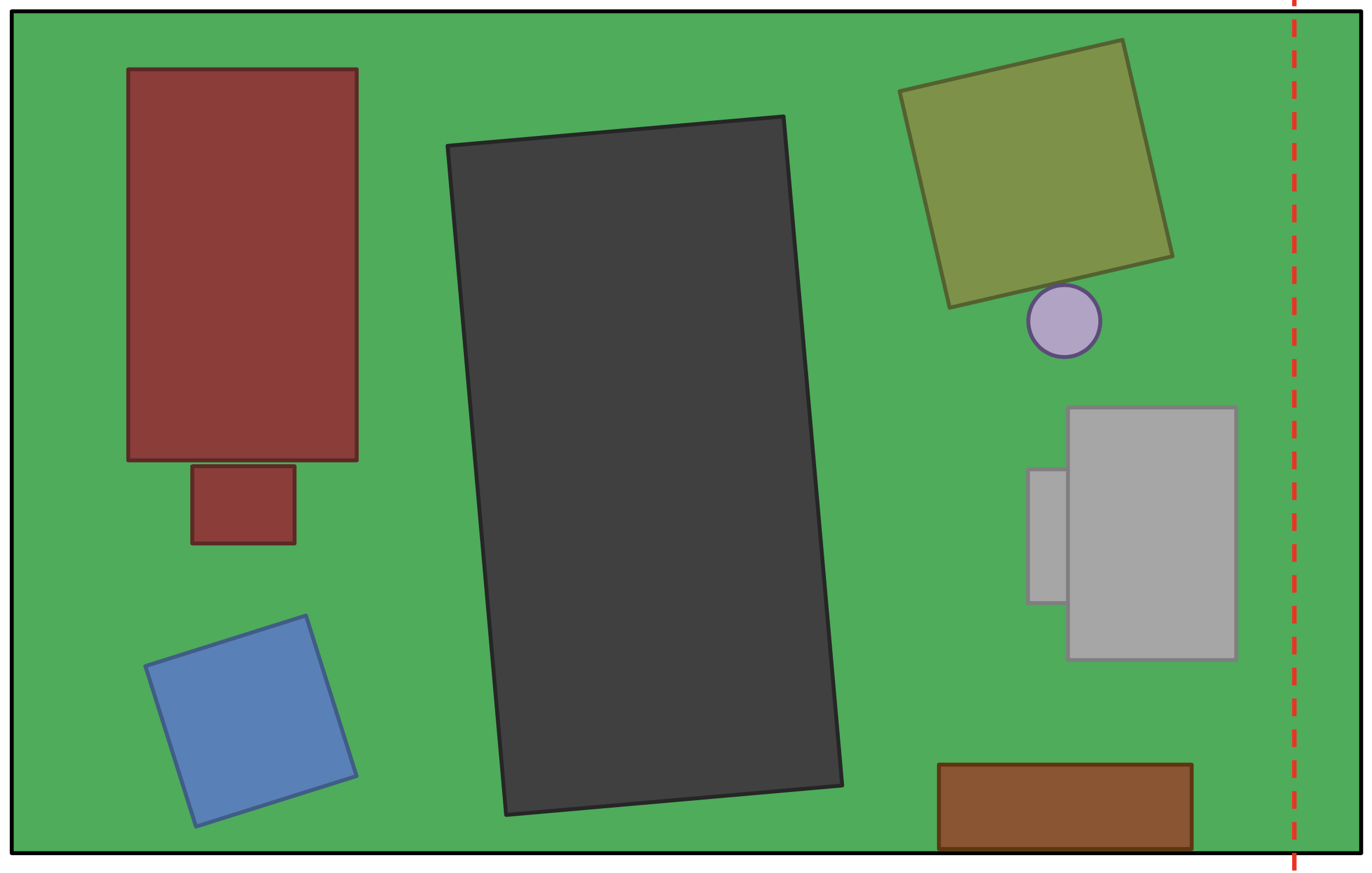}
\end{subfigure}

\begin{subfigure}[b]{\linewidth}
\includegraphics[width=\linewidth]{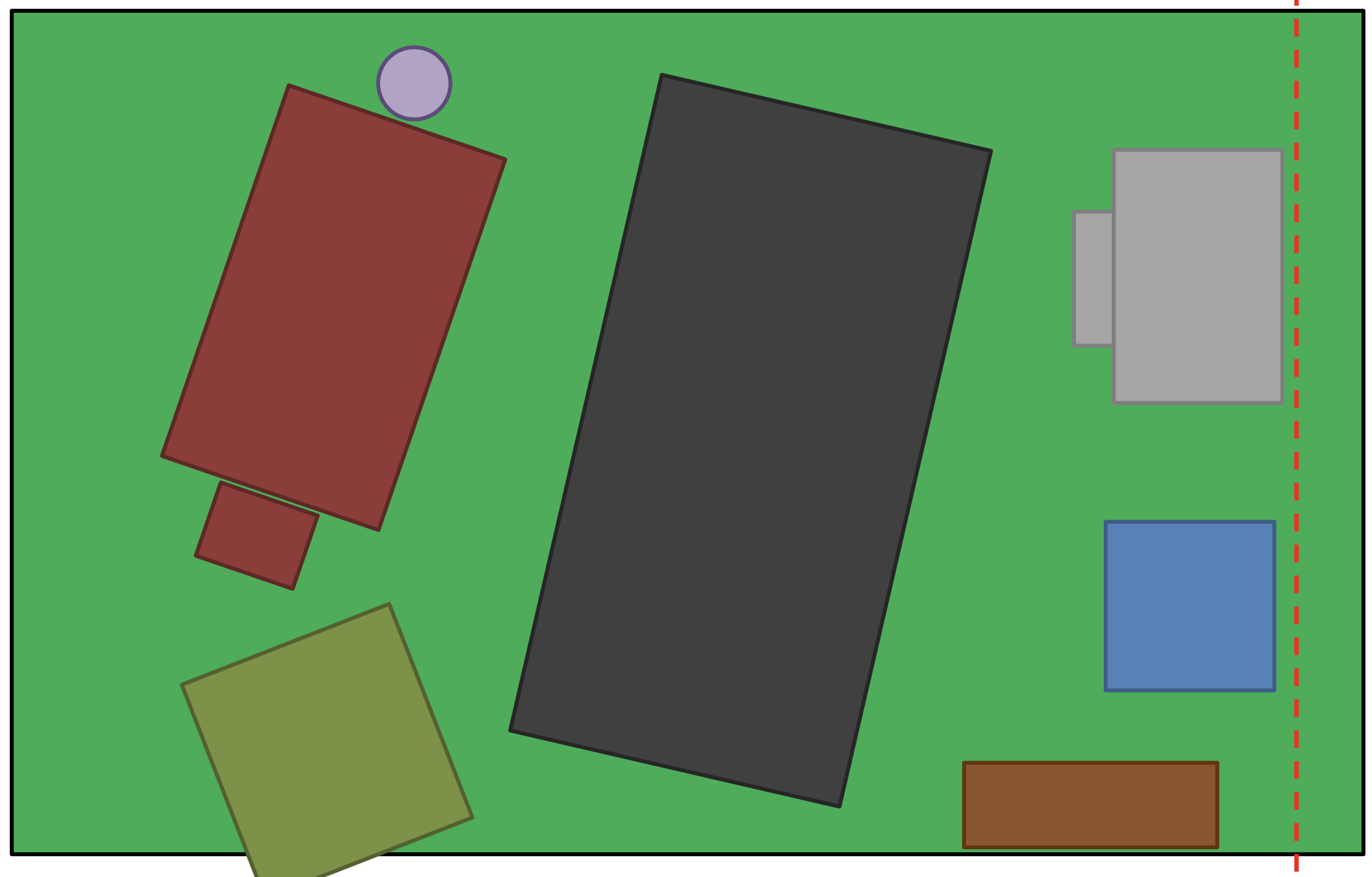}
\end{subfigure}

\begin{subfigure}[b]{\linewidth}
\includegraphics[width=\linewidth]{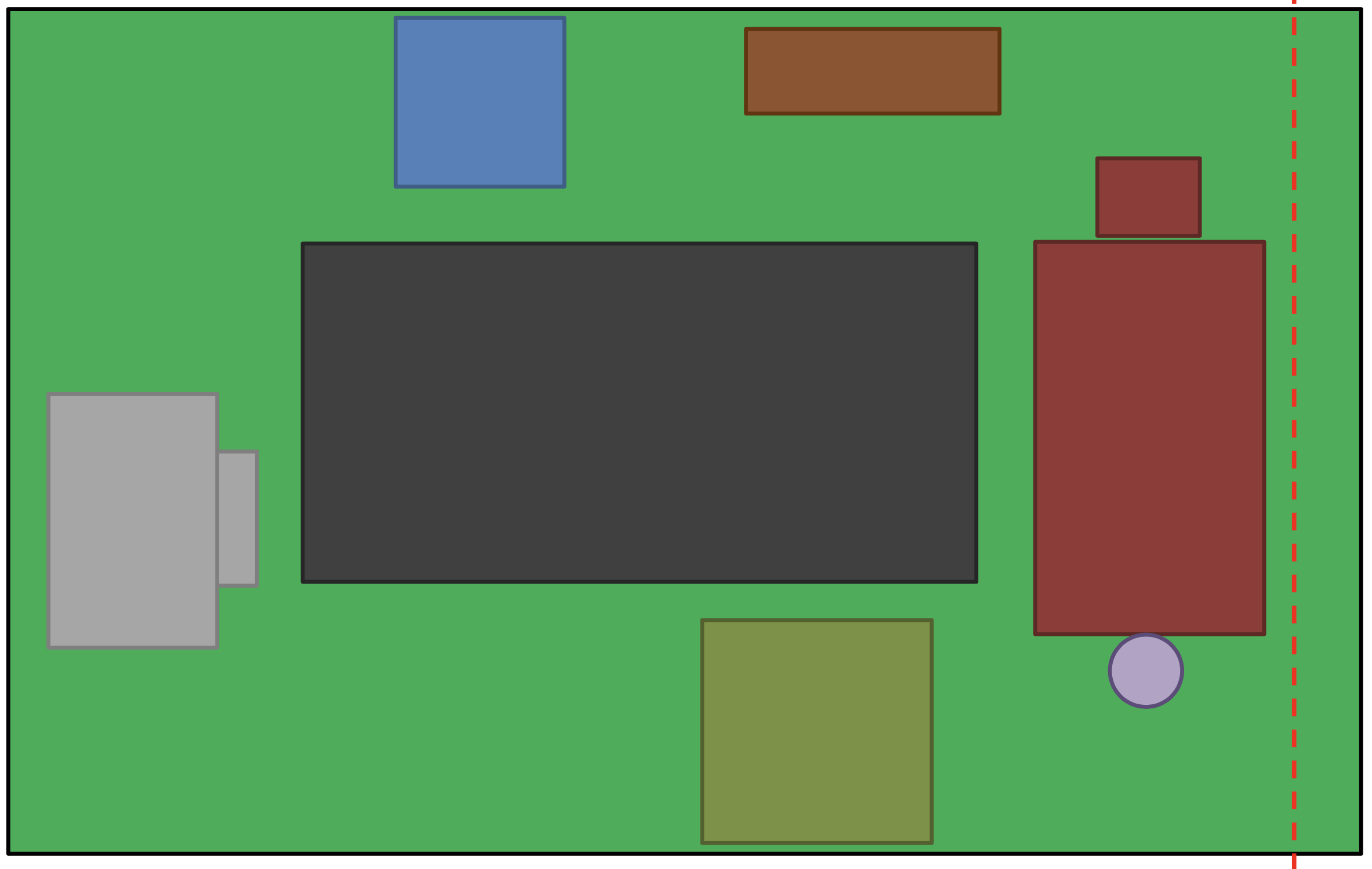}
\end{subfigure}

\caption{Additional possible room furniture layouts used in study protocol. Furniture pieces are as follows; \textbf{black} - mat; \textbf{brown} - shelf; \textbf{olive} - chair; \textbf{grey} - desk + chair; \textbf{red}blue - table; \textbf{red} - sofa + end table; \textbf{purple} - lamp.}
\label{fig:activity_2}
\end{figure}

\EOD
\end{document}